\documentclass{article}

\usepackage[a4paper, total={6.5in, 8.5in}]{geometry}
\usepackage{graphicx}
\usepackage{authblk}
\usepackage{soul}
\usepackage{amsmath}
\usepackage[
backend=biber,
giveninits=true,
style=numeric-comp,
sorting=none,
maxcitenames=1,
natbib=true,
sortlocale=en_US,
url=false, 
doi=true,
eprint=false
]{biblatex}
\usepackage{geometry}
\usepackage{atbegshi}
\usepackage{float,lscape}
\usepackage{multirow}
\usepackage{blindtext}
\usepackage{amssymb}
\usepackage{ragged2e}
\usepackage{blindtext}
\usepackage{longtable}
\usepackage{comment}
\usepackage{graphicx}
\usepackage{xcolor}
\usepackage{caption}
\usepackage{amsmath}
\usepackage{bm}
\usepackage{changepage}

\usepackage{amsfonts}
\usepackage{tabularx}
\usepackage{lipsum}
\usepackage[title]{appendix}
\usepackage{mathtools}
\usepackage{circuitikz}
\usetikzlibrary{patterns}
\usepackage{hyperref}
\usepackage{caption}
\usepackage{siunitx}

\hypersetup{
	colorlinks   = true, 
	urlcolor     = blue, 
	linkcolor    = blue, 
	citecolor   = red 
}
\providecommand{\keywords}[1]
{
  \small	
  \textbf{\textit{Keywords---}} #1
}

\title{Applications of Scientific Machine Learning for the Analysis of Functionally Graded Porous Beams}

\author[1]{Mohammad Sadegh Eshaghi}
\author[3]{Mostafa Bamdad}
\author[3]{Cosmin Anitescu}
\author[3,4]{Yizheng Wang}
\author[1,2]{Xiaoying Zhuang\thanks{Corresponding author. Email: zhuang@iop.uni-hannover.de}}
\author[3]{Timon Rabczuk}

\affil[1]{Chair of Computational Science and Simulation Technology, Institute of Photonics, Department of Mathematics and Physics, Leibniz University Hannover, 30167 Hannover, Germany}
\affil[2]{Department of Geotechnical Engineering, College of Civil Engineering, Tongji University, Siping Road 1239, 200092 Shanghai, China}
\affil[3]{Institute of Structural Mechanics, Bauhaus-Universität Weimar, Germany}
\affil[4]{Department of Engineering Mechanics, Tsinghua University, Beijing, China}

\date{}

\begin{document}

\maketitle

\begin{center}
    \vspace{1cm}
    \textbf{\Large Preprint Notice} \\
    \vspace{0.5cm}
    This is a preprint version of the article. \\
    The final version is published in \emph{Neurocomputing}. \\
    Volume 619, Page 129119, February 28, 2025. \\
    DOI: \href{https://doi.org/10.1016/j.neucom.2024.129119}{https://doi.org/10.1016/j.neucom.2024.129119} \\
    \href{https://www.sciencedirect.com/science/article/pii/S0925231224018903}{https://www.sciencedirect.com/science/article/pii/S0925231224018903} \\
    \vspace{0.5cm}
    \textbf{Disclaimer:} The preprint may differ from the final version published by the journal.
    \vspace{2cm}
\end{center}

\begin{abstract}
This study investigates different Scientific Machine Learning (SciML) approaches for the analysis of functionally graded (FG) porous beams and compares them under a new framework. The beam material properties are assumed to vary as an arbitrary continuous function. The methods consider the output of a neural network/operator as an approximation to the displacement fields and derive the equations governing beam behavior based on the continuum formulation. The methods are implemented in the framework and formulated by three approaches: (a) the vector approach leads to a Physics-Informed Neural Network (PINN), (b) the energy approach brings about the Deep Energy Method (DEM), and (c) the data-driven approach, which results in a class of Neural Operator methods. Finally, a neural operator has been trained to predict the response of the porous beam with functionally graded material under any porosity distribution pattern and any arbitrary traction condition. The results are validated with analytical and numerical reference solutions. The data and code accompanying this manuscript will be publicly available at \href{https://github.com/eshaghi-ms/DeepNetBeam}{https://github.com/eshaghi-ms/DeepNetBeam}.
\end{abstract}


\section{Introduction}

In the field of engineering, exploring novel materials with customized qualities has led to the emergence of Functionally Graded Materials (FGMs) which represent a class of composite materials characterized by gradual variations in material properties, resulting in unique characteristics that are not possible with homogeneous materials \cite{li2020review}. This continuous variation offers a range of advantages over homogeneous materials, including the ability to customize properties such as mechanical, thermal, electrical, and magnetic characteristics. Moreover, this refinement reduces thermal stress through gradual variation of thermal expansion coefficients, thereby minimizing the risk of thermal cracking \cite{audouard2024resistance}. FGMs also provide thermal protection and corrosion resistance, making them valuable in the aerospace, automotive, and energy industries. Their lightweight design, combined with structural integrity, allows the creation of structures suitable for weight-sensitive applications \cite{putz2021microstructure}.

On the other hand, due to the broad application of beams across diverse fields including civil engineering, marine, military, and aeronautics industries, it is imperative to investigate the mechanical response of beams under different materials and loading conditions. Addressing this demand, beam theories are used to simplify the analysis of complex structures and predict their behavior. The primary goals of various beam theories, such as Classical Beam Theory (CBT), First Order Shear Deformation Theory (FSDT), and Higher-Order Shear Deformation Theory (HSDT), are to reach a balance between precision and efficiency \cite{ochsner2021euler}. The fundamental differences between these theories stem from their methods of modeling beam behavior. CBT simplifies the analysis by assuming that the cross-section remains plane and perpendicular to the beam's axis after deformation, neglecting shear deformations. FSDT considers shear deformation effects but assumes constant transverse shear strains across the beam's thickness. In contrast, HSDT accounts for variable transverse shear strains, resulting in a more accurate shear deformation. Significant research has been carried out in this area \cite{mohammadimehr2018bending, gatheeshgar2020optimised}. Therefore it is of interest to investigate the analysis of FGMs for beams. Numerous studies in the literature also explore the mechanical analysis of FG beams, as evidenced by references \cite{babaei2022functionally, ramteke2023computational, chen2016free, kiarasi2021review, chen2023functionally, agarwal2006large}.


While beam theories have their utility in specific scenarios, their effectiveness might be limited when compared to the flexibility, autonomy, and data-driven nature of Machine Learning (ML) approaches in computational mechanics \cite{fallah2023physics, turan2023free}. ML has the potential to revolutionize the field by offering more accurate solutions, particularly for complex and real-world problems \cite{nguyen2021parametric, samaniego2020energy}. For instance, Ebrahimi and Ezzati \cite{ebrahimi2023machine} studied Young's modulus estimation in functionalized graphene-reinforced nanocomposites using ML. Their findings recommend implementing these models for faster results in engineering applications. Additionally, the research by Ahmed et al. \cite{ahmed2023prediction} applied ML models (ANN, XGBoost, SVM) to predict shear behavior in ultra-high performance concrete I-shaped beans, where XGBoost exhibited higher accuracy. It highlights the improved accuracy of ML models compared to conventional design methods. Taking advantage of Artificial Neural Networks (ANNs), Mojtabaei et al. \cite{mojtabaei2023predicting} successfully predicted elastic critical buckling loads and modal decompositions in thin-walled structural elements, showing their ability to address complicated challenges.

ML finds a wide variety of applications due to its ability to learn patterns from data and then generate the expected output. One significant application of this approach can be found in solving Partial Differential Equations (PDEs) \cite{blechschmidt2021three, aarts2001neural, dockhorn2019discussion}. The potential of using deep learning methods to approximate solutions for PDEs was investigated by Beck et al. \cite{beck2020overview}. These methods provide new possibilities for solving PDEs in various fields, especially in complex system modeling. The readers can find the history of different ANN methods for solving differential equations in the book by Yadav et al. \cite{yadav2015introduction}. This ML application led to the development of the field of Scientific Machine Learning (SciML). 

SciML is an interdisciplinary field that combines techniques from scientific computing, machine learning, and domain-specific sciences. Its main purpose is to develop algorithms that combine the knowledge of physical systems (e.g., differential equations, and conservation laws) and the power of ML \cite{cuomo2022scientific}. In traditional scientific computing, researchers use mathematical models to study the behavior of systems. However, these models often depend on simplifications and assumptions that may not capture the complexity of real-world phenomena. On the other hand, ML approaches are suitable for learning patterns and making predictions from data but may lack interpretability and struggle with extrapolation outside of the data distribution. SciML bridges this gap by merging scientific knowledge into ML algorithms \cite{hey2020machine}.

Another utilization of SciML is in addressing beam-related problems within the field of structural mechanics. Beam-related problems can benefit from SciML's potential to improve their analysis, design, and performance evaluation across diverse loading scenarios \cite{bazmara2023application}. Therefore, we aim to formulate a new framework, named DeepNetBeam (DNB), which uses SciML for analyzing FG beams. For this purpose, there are three possible approaches to consider: (a) the vector approach followed in continuum mechanics, which leads to Physics-Informed Neural Networks (PINN), (b) the energy approach based on the principle of minimum total potential energy, leading to Deep Energy methods (DEM), or (c) the data-driven approach, resulting in Neural Operator methods. Therefore, the novelty of the current work lies in demonstrating the use of Neural Operators for beam problems for the first time and unifying different scientific machine learning methods under one framework.

PINNs in the area of computational science are attracting attention for their role in solving PDEs. These neural networks are specifically designed to learn and predict the behavior of physical systems while following the fundamental laws of physics. By incorporating domain knowledge and physical equations, PINNs enable accurate predictions even with limited data \cite{haghighat2021physics, anitescu2023physics}. Kapoor et al. \cite{taniya2023physics} investigated the application of PINN in simulating complex beam systems, showing that the relative error in computing beam displacement remains low as model complexity increases. The article also demonstrates the algorithm's effectiveness in solving inverse problems and discovering force functions and model parameters. Moreover, the study by Roy et al. \cite{roy2023deep}, introduces a deep learning framework that uses PINNs to deal with linear elasticity problems. This approach combines data-driven deep learning with traditional numerical techniques, covering a range of mechanics and material science problems. Utilizing variational formulation as the loss function, the approach demonstrated good accuracy in some examples, suggesting its potential as a low-fidelity surrogate for applications like reliability analysis and design optimization. Goswami et al. \cite{goswami2020transfer} introduced a PINN algorithm based on transfer learning for predicting crack paths in fracture mechanics. In addition, Fallah and Aghdam \cite{fallah2023physics} proposed a PINN approach to analyze bending and free vibration in TDFG porous beams on an elastic foundation. The research explores the impacts of material distribution, porosity, and foundation on beam behavior, showcasing the flexibility and usefulness of the proposed PINN in predicting the behavior of TDFG beams. As another example, Luong et al. \cite{luong4057311deep} used a parallel network in developing a PINN framework for solving beam bending problems by converting higher-order PDEs into lower-order ones. 

The second approach, DEM, introduced by Samaniego et al. \citep{samaniego2020energy}, differs from the PINN framework by centering its optimization process around minimizing the potential energy of physical systems rather than solely focusing on PDE residuals. While DEM is applicable to physical systems that adhere to the principle of minimum potential energy, it was quickly applied in various fields \citep{nguyen2020deep, gao2022dhem, fuhg2022mixed, matsubara2020deep, zhuang2021deep, abueidda2023enhanced, he2023use, he2023deep}. Nonetheless, DEM implementation in analyzing FG beams is quite rare, and existing works do not have a detailed analysis of robustness or computational efficiency \cite{mojahedin2021deep, mojahedin2022deep}. 

Finally, Neural Operators are a class of models designed to learn mappings between function spaces in order to solve PDEs directly from data. While PINNs and DEMs learn the solution of a single instance of a boundary value problem at a time, Neural Operators learn the mapping between the input data and the solution fields for a particular PDE on a fixed geometry \cite{kovachki2023neural}. Implementing neural operators for FG porous beams is pursued as the third approach within the DNB framework.

The current study investigates the different machine-learning based approaches for the analysis of porous beams with functionally graded materials. In fact, in the DNB framework, by considering the output of a neural network as an approximation to the displacement fields and deriving the formulation for equations governing beam behavior, the focus is on developing a more adaptable approach to beam analysis. DNB  further endeavors to overcome the limitations resulting from assumptions in traditional beam theories, ultimately improving the accuracy of beam response estimations. Through solving some problems, such as parametric investigations on a porous beam, considering factors such as the slenderness ratio, porosity coefficient, and distribution type, the accuracy and effectiveness of the proposed DNB in the context of FG beam analysis are investigated. Additionally, a neural network has been trained to predict the response of an FG porous beam under any porosity distribution pattern and arbitrary traction condition. By doing this, the research contributes to and takes advantage of the capabilities of SciML to provide physics-informed and data-driven paradigms for FG porous beam analysis and the prediction of structural behavior.

The following organized structure is adopted to facilitate the comprehensive exploration of DNB in this research: In section \ref{sec:Formulation}, we provide the formulation of DNB, detailing the integration of SciML and equations governing beam behavior. Section \ref{sec:Result} exemplifies DNB's application through different problems and a parametric study is presented in section \ref{sec:ParametricAnalysis}. The conclusion in Section \ref{sec:Conclusion} summarizes the paper's contributions, underlining DNB's potential to enhance structural analysis through SciML capabilities.

\section{Formulation} \label{sec:Formulation}
In general, beam theories have been developed using one of the following two approaches \cite{reddy2022theories}:
\begin{enumerate}

\item Theories relying on the presumed displacement expansions in terms of the powers of the thickness coordinate and unknown functions:
\begin{equation}
u_i(x, y, z, t) = \sum_{j=0}^m (z)^j \varphi_i^{(j)}, \quad i=1,2,3
\end{equation}
where \(u_i (i=1,2,3)\) are total displacement of a point (x,y,z) in the body and \(\varphi_i^{(j)}\) are the displacement functions to be determined. 

\item Theories relying on the presumed stress expansions: 
\begin{equation}
\sigma_i(x, y, z, t) = \sum_{j=0}^m (z)^j \psi_i^{(j)}, \quad i=1,2,\dots,6
\end{equation}
where \(\sigma_i\) are the stress components \(\sigma_1 = \sigma_{xx}\), \(\sigma_2 = \sigma_{yy}\), \(\sigma_3 = \sigma_{zz}\), \(\sigma_4 = \sigma_{yz}\), \(\sigma_5 = \sigma_{xz}\), \(\sigma_6 = \sigma_{xy}\), and \(\psi_i^{(j)}\) are the stress functions to be determined.
\end{enumerate}

In this paper, we define our coordinate system as follows: The \(x\)-axis runs along the length of the beam and passes through its geometric centroid. The \(z\)-axis is oriented upward, perpendicular to the beam's axis, and the \(y\)-axis is directed outward perpendicular to the \(xz\)-plane. Our current focus is on bending about the \(y\)-axis, and we are developing the framework using displacement expansions. However, it's important to note that this formulation can be extended to stress expansions as well.

Let the displacement vector be denoted as \(\mathbf{u} = u_x \hat{\mathbf{e}}_x + u_y \hat{\mathbf{e}}_y + u_z \hat{\mathbf{e}}_z\) with \((u_x,u_y,u_z)\) being the component referred to the \((x,y,z)\) coordinates. For bending in the \(xz\)-plane (i.e, bending about \(y\) axis) the displacement field is:

\begin{equation}
\mathbf{u}(x, z) = \mathcal{N}_x \hat{\mathbf{e}}_x  + \mathcal{N}_z \hat{\mathbf{e}}_z
\end{equation}
where \(\mathcal{N}_x\) and \(\mathcal{N}_z\) are the outputs of a Deep Neural Network, \(\mathcal{N}\), which is defined as follows: 

\begin{equation}
\mathcal{N}(x,z;\theta) = [\mathcal{N}_x, \mathcal{N}_z] = f^{[L]}(f^{[L-1]}(\cdots f^{[l]}(\cdots f^{[1]}(x,z))))
\end{equation}
where \(L\) is the number of network layers, including an input layer, hidden layers, and an output layer, \(\theta\) is network parameters, and \(f^{[l]}\) represents the \(l\)-th layer of networks as follows:
\begin{equation}
f^{[1]} = \mathcal{A}^{[1]}\left(\mathbf{W}^{[1]}\divideontimes [x,z]^\top+\mathbf{b}^{[1]}\right) 
\end{equation}
\begin{equation}
f^{[l]} = \mathcal{A}^{[l]}\left(\mathbf{W}^{[l]}\divideontimes f^{[l-1]}+\mathbf{b}^{[l]}\right) , \quad l=2,3,\dots,L 
\end{equation}
where the weights and biases (network parameters, \(\theta\)) of each layer are represented by \(\mathbf{W}^{[l]}\) and \(\mathbf{b}^{[l]}\) respectively, for the \(l\)-th layer, \(\mathcal{A}^{[l]}\) is the activation function, and the symbol \(\divideontimes\) is matrix multiplication \(\left( \cdot \right)\) for fully connected layers and convolution operation \(\left( \ast \right)\) for convolutional layers. Therefore, we can rewrite the displacement field for the beam:
\begin{equation}
\begin{aligned}
& \mathbf{u}_x(x, z) = \mathcal{N}_x (x,z;\theta) \\
& \mathbf{u}_y(x, z) = 0 \\
& \mathbf{u}_z(x, z) = \mathcal{N}_z (x,z;\theta) \label{eq:displacement_field}
\end{aligned}
\end{equation}

In addition, the measure of strain in solid mechanics is the Green-Lagrange strain tensor defined by \cite{reddy2013introduction}
\begin{equation}
\mathbf{E} = \frac{1}{2} \biggl[ \nabla \mathbf{u} + (\nabla \mathbf{u})^\top + (\nabla \mathbf{u}) \cdot (\nabla \mathbf{u})^\top \biggr] \label{eq:Green_Lagrange_strain_tensor}
\end{equation}

where \(\nabla\) is the gradient operator with respect to material coordinates \(\mathbf{X}\) in the reference configuration:
\begin{equation}
\nabla = \hat{\mathbf{E}}_1 \frac{\partial}{\partial X_1} + \hat{\mathbf{E}}_2 \frac{\partial}{\partial X_2} + \hat{\mathbf{E}}_3 \frac{\partial}{\partial X_3} = \hat{\mathbf{E}}_i \frac{\partial}{\partial X_i} \label{eq:nabla_MaterialCoordinates}
\end{equation}

where \((\hat{\mathbf{E}}_1, \hat{\mathbf{E}}_2, \hat{\mathbf{E}}_3)\) are the unit base vectors in the coordinate system \((X_1, X_2, X_3)\). In the analysis of beams, the deformations are relatively small (\(X_i \approx x_i\)) but there are significant rotations about the \(y\)-axis. This means that the squares and products of \(\partial u_z/\partial x\) and \(\partial u_z/\partial y\) cannot be ignored, however, squares and products of \(\partial u_x/\partial x\), \(\partial u_y/\partial y\), and \(\partial u_z/\partial z\) can be considered negligible in this context. The strains, which are a result of the components of the Green strain tensor, are referred to as Föppl-von Kármán strains \cite{lewicka2011foppl}:
\begin{align}
& \varepsilon_{xx} = \frac{\partial u_x}{\partial x} + \frac{1}{2} \left( \frac{\partial u_z}{\partial x} \right)^2, \quad
\varepsilon_{yy} = \frac{\partial u_y}{\partial y} + \frac{1}{2} \left( \frac{\partial u_z}{\partial y} \right)^2, \quad
\varepsilon_{zz} = \frac{\partial u_z}{\partial z},  \\
& 2\varepsilon_{xy} = \frac{\partial u_x}{\partial y} + \frac{\partial u_y}{\partial x} + \frac{\partial u_z}{\partial x} \frac{\partial u_z}{\partial y}, \quad
2\varepsilon_{xz} = \frac{\partial u_x}{\partial z} + \frac{\partial u_z}{\partial x},\quad
2\varepsilon_{yz} = \frac{\partial u_y}{\partial z} + \frac{\partial u_z}{\partial y}. 
\end{align}

Therefore, the nonzero strain tensor components referred to the rectangular Cartesian system \((x,y,z)\), associated with the displacement field in Eq. \ref{eq:displacement_field} are obtained:
\begin{align}
\varepsilon_{xx} = \frac{\partial \mathcal{N}_x}{\partial x} + \frac{1}{2} \left( \frac{\partial \mathcal{N}_z}{\partial x} \right)^2, \quad
\varepsilon_{zz} = \frac{\partial \mathcal{N}_z}{\partial z}, \quad
\varepsilon_{xz} = \frac{1}{2} \left( \frac{\partial \mathcal{N}_x}{\partial z} +  \frac{\partial \mathcal{N}_z}{\partial x} \right). \label{eq:nonzero_strain_tensor}
\end{align}


The only nonzero components of the rotation vector \(\bm{\omega}\) and the curvature tensor \(\bm{\chi}\) associated with the displacement field in Eq. \ref{eq:displacement_field} are presented as:
\begin{equation}
\omega_y = \frac{1}{2}\left( \frac{\partial{\mathcal{N}_x}}{\partial{z}} - \frac{\partial{\mathcal{N}_z}}{\partial{x}} \right) \label{eq:rotation_vector}
\end{equation}
\begin{equation}
\chi_{xy} = \frac{1}{2} \frac{\partial{\omega_y}}{\partial{x}} = \frac{1}{4} \left( \frac{\partial^2{\mathcal{N}_x}}{\partial{x}\partial{z}} - \frac{\partial^2{\mathcal{N}_z}}{\partial{x^2}} \right)
\end{equation}
\begin{equation}
\chi_{yz} = \frac{1}{2} \frac{\partial{\omega_y}}{\partial{z}} = \frac{1}{4} \left( \frac{\partial^2{\mathcal{N}_x}}{\partial{z^2}} - \frac{\partial^2{\mathcal{N}_z}}{\partial{x}\partial{z}} \right) 
\end{equation}

However, in the case of an isotropic, linear elastic material, the stress-strain relations in three dimensions can be expressed as follows:
\begin{equation}
\sigma_{ij} = 2\mu\varepsilon_{ij}+\lambda\delta_{ij}\varepsilon_{kk} \label{eq:Constitutive}
\end{equation}
Here, the parameters \(\mu\) and \(\lambda\) are known as the Lame parameters and they are defined as \cite{sadd2009elasticity}:
\begin{equation}
\lambda = \frac{E\nu}{(1+\nu)(1-2\nu)}, \quad \mu = \frac{E}{2(1+\nu)} \label{eq:Lame_parameters}
\end{equation}
with \(E\) represents Young's modulus and \(\nu\) represents Poisson's ration. Therefore, the constitutive relations in Eq. \ref{eq:Constitutive} can be simplified as follows:

\begin{align}
\label{eq:sigm_xx} &&&&&&&&&&& \sigma_{xx}  = &(2\mu+\lambda) \frac{\partial{\mathcal{N}_x}}{\partial{x}} &+& \lambda \frac{\partial{\mathcal{N}_z}}{\partial{z}} &+& (\mu + \frac{\lambda}{2}) \left( \frac{\partial \mathcal{N}_z}{\partial x} \right)^2 &&&&&&&&&&&\\ 
\label{eq:sigm_yy} &&&&&&&&&&& \sigma_{yy}  = &\lambda \frac{\partial{\mathcal{N}_x}}{\partial{x}} &+& \lambda \frac{\partial{\mathcal{N}_z}}{\partial{z}} &+ & \frac{\lambda}{2} \left( \frac{\partial \mathcal{N}_z}{\partial x} \right)^2 &&&&&&&&&&& \\ 
\label{eq:sigm_zz} &&&&&&&&&&& \sigma_{zz}  = &\lambda \frac{\partial{\mathcal{N}_x}}{\partial{x}} &+& (2\mu+\lambda) \frac{\partial{\mathcal{N}_z}}{\partial{z}} &+ & \frac{\lambda}{2} \left( \frac{\partial \mathcal{N}_z}{\partial x} \right)^2 &&&&&&&&&&& 
\end{align}
\begin{equation}
\sigma_{xz}  =\mu \left( \frac{\partial \mathcal{N}_x}{\partial z} + \frac{\partial \mathcal{N}_z}{\partial x} \right) \label{eq:sigm_xz}
\end{equation}

Therefore, the displacement fields in Eq. \ref{eq:displacement_field} involve two unknowns, specifically denoted as \(\mathcal{N}_{x}\) and \(\mathcal{N}_{z}\), which we need to solve for. In turn, these are determined by the network parameters \(\mathbf{W}\) and \(\mathbf{b}\). In deriving the governing equations, as mentioned earlier, there are three alternative approaches to consider: (a) the vector approach followed in continuum mechanics, which here leads to Physics-Informed Neural Networks, (b) the energy approach based on the principle of minimum total potential energy, leading to Deep Energy methods, or (c) the data-driven approach, resulting in Neural Operator methods.

The vector approach typically involves defining suitable forces acting on an infinitesimal element taken from the continuum, while the energy approach focuses on constructing the energy functional for the system. However, the data-driven approach solves the problem without explicitly considering the underlying physical laws. Moreover, it is worth mentioning that there are also formulations based on Hellinger-Reisner \cite{weissman1990unified} and Ho-Washizo \cite{he1997equivalent} principles, which are not discussed in this article and can be the subject of future studies.

\subsection{Vector Approach - Physics-Informed Neural Networks} \label{sec:PINNs}
The fundamental principle of linear momentum balance, when applied to a deformed solid continuum and expressed in terms of the second Piola-Kirchhoff stress tensor \(\mathbf{S}\) is given by the equation:
\begin{equation}
\mathbf{\nabla} \cdot \left[ \mathbf{S} \cdot \left( \mathbf{I} + \mathbf{\nabla u} \right) \right] + \hat{\mathbf{f}} = 0 \label{eq:Piola-Kirchhoff}
\end{equation}
where \(\mathbf{\nabla}\) represents the gradient operator with respect to the material coordinate \(\mathbf{X}\) and \(\hat{\mathbf{f}}\) represents the body force per unit volume in the undeformed body. Therefore, when we expand the vector form of the equations of motion for a 2-D solid continuum in the rectangular Cartesian coordinate system (x, y, z), we get the following equations:
\begin{equation}
\frac{\partial}{\partial{x}} 
\left( \sigma_{xx} + \frac{\partial{u_x}}{\partial{x}} \sigma_{xx} + \frac{\partial{u_x}}{\partial{z}} \sigma_{xz}  \right) 
+
\frac{\partial}{\partial{z}} 
\left( \sigma_{xz} + \frac{\partial{u_x}}{\partial{x}} \sigma_{xz} + \frac{\partial{u_x}}{\partial{z}} \sigma_{zz}  \right) 
+ 
f_x(x,z)
= 0 \label{eq:vectorform-1}
\end{equation}

\begin{equation}
\frac{\partial}{\partial{x}} 
\left( \sigma_{xz} + \frac{\partial{u_z}}{\partial{x}} \sigma_{xx} + \frac{\partial{u_z}}{\partial{z}} \sigma_{xz}  \right) 
+
\frac{\partial}{\partial{z}} 
\left( \sigma_{zz} + \frac{\partial{u_z}}{\partial{x}} \sigma_{xz} + \frac{\partial{u_z}}{\partial{z}} \sigma_{zz}  \right) 
+ 
f_z(x,z)
= 0 \label{eq:vectorform-2}
\end{equation}

Hence, the loss function can be defined as follows by incorporating Eqs. \ref{eq:sigm_xx}-\ref{eq:sigm_xz} into Eqs. \ref{eq:vectorform-1} and \ref{eq:vectorform-2}, while also taking into consideration that Eqs. \ref{eq:vectorform-1} and \ref{eq:vectorform-2} must hold throughout the entire domain.

\begin{equation}
\mathcal{L}(\mathbf{W}, \mathbf{b}) = \frac{1}{N_d} \, \sum_{i=1}^{N_d} \left[ \left(\frac{\partial \mathcal{M}_{xx}^i}{ \partial x} + \frac{\partial \mathcal{M}_{xz}^i}{ \partial z} + f_x^i\right)^2 + \left( \frac{\partial \mathcal{M}_{zx}^i}{ \partial x} + \frac{\partial \mathcal{M}_{zz}^i}{ \partial z} + f_z^i\right)^2 \right]
\end{equation}
In this equation, \(N_d\) specifies the total number of collocation points within the entire domain, \(f^i\) and \(\mathcal{M}^i\) represent  \(f(x_i,z_i)\) and \(\mathcal{M}(x_i,z_i)\), respectively, and \(\mathcal{M}_{pq}(x_i,z_i)\) is defined as follows:

\begin{align}
\begin{split}
\mathcal{M}_{xx}(x_i,z_i) = \,
& (2\mu+\lambda)\frac{\partial{\mathcal{N}_x}}{\partial{x}}+\lambda \frac{\partial{\mathcal{N}_z}}{\partial{z}}
 +
 (2\mu+\lambda) \left( \frac{\partial{\mathcal{N}_x}}{\partial{x}} \right)^2 + \mu \left( \frac{\partial{\mathcal{N}_x}}{\partial{z}} \right)^2 + (\mu+\frac{\lambda}{2}) \left( \frac{\partial{\mathcal{N}_z}}{\partial{x}} \right)^2 \\
+ \,
& \lambda\frac{\partial{\mathcal{N}_x}}{\partial{x}}\frac{\partial{\mathcal{N}_z}}{\partial{z}} + \mu\frac{\partial{\mathcal{N}_x}}{\partial{z}}\frac{\partial{\mathcal{N}_z}}{\partial{x}}
+ 
(\mu+\frac{\lambda}{2}) \frac{\partial{\mathcal{N}_x}}{\partial{x}} \left( \frac{\partial{\mathcal{N}_z}}{\partial{x}} \right)^2 \label{eq:M_xx}
\end{split}
\\
\begin{split}
\mathcal{M}_{xz}(x_i,z_i) = \,
&\mu \frac{\partial{\mathcal{N}_x}}{\partial{z}}
+\mu \frac{\partial{\mathcal{N}_z}}{\partial{x}}
+
(\mu+\lambda) \frac{\partial{\mathcal{N}_x}}{\partial{x}} \frac{\partial{\mathcal{N}_x}}{\partial{z}}  
+ 
\mu \frac{\partial{\mathcal{N}_x}}{\partial{x}} \frac{\partial{\mathcal{N}_z}}{\partial{x}}
+ (2\mu+\lambda)  \frac{\partial{\mathcal{N}_z}}{\partial{z}}  \frac{\partial{\mathcal{N}_x}}{\partial{z}}
\\
+ \, &
\frac{\lambda}{2} \frac{\partial{\mathcal{N}_x}}{\partial{z}} \left( \frac{\partial{\mathcal{N}_z}}{\partial{x}} \right)^2
\end{split}
\\
\begin{split}
\mathcal{M}_{zx}(x_i,z_i) = \,
&\mu \frac{\partial{\mathcal{N}_x}}{\partial{z}}
+\mu \frac{\partial{\mathcal{N}_z}}{\partial{x}}
+
(\mu+\lambda) \frac{\partial{\mathcal{N}_z}}{\partial{z}} \frac{\partial{\mathcal{N}_z}}{\partial{x}}  
+ 
\mu \frac{\partial{\mathcal{N}_x}}{\partial{z}} \frac{\partial{\mathcal{N}_z}}{\partial{z}}
+ (2\mu+\lambda)  \frac{\partial{\mathcal{N}_x}}{\partial{x}}  \frac{\partial{\mathcal{N}_z}}{\partial{x}}
\\
+ \, &
(2\mu+\frac{\lambda}{2}) \left( \frac{\partial{\mathcal{N}_z}}{\partial{x}} \right)^3
\end{split}
\\ 
\begin{split}
\mathcal{M}_{zz}(x_i,z_i) = \,
& (2\mu+\lambda)\frac{\partial{\mathcal{N}_z}}{\partial{z}}+\lambda \frac{\partial{\mathcal{N}_x}}{\partial{x}}
 +
 (2\mu+\lambda) \left( \frac{\partial{\mathcal{N}_z}}{\partial{z}} \right)^2 
 + (\mu+\frac{\lambda}{2}) \left( \frac{\partial{\mathcal{N}_z}}{\partial{x}} \right)^2 \\
+ \,
& \lambda\frac{\partial{\mathcal{N}_x}}{\partial{x}}\frac{\partial{\mathcal{N}_z}}{\partial{z}} 
+ \mu\frac{\partial{\mathcal{N}_x}}{\partial{z}}\frac{\partial{\mathcal{N}_z}}{\partial{x}}
+ 
\frac{\lambda}{2} \frac{\partial{\mathcal{N}_z}}{\partial{z}} \left( \frac{\partial{\mathcal{N}_z}}{\partial{x}} \right)^2
\end{split}
\end{align}
Now we've formulated all the necessary equations to determine \(\mathbf{W}\) and \(\mathbf{b}\), so that
\begin{equation}
    \left( \mathbf{W}^*, \mathbf{b}^* \right) = \operatorname*{argmin}_{\mathbf{W}, \mathbf{b}} \mathcal{L}(\mathbf{W}, \mathbf{b}) \label{eq:gradient_descent_1}
\end{equation}
and for achieving the values of \(\mathbf{W}^*, \mathbf{b}^*\), different methods such as gradient descent method \cite{ruder2016overview} can be used, as follows:
\begin{equation}
    W_{j+1} \leftarrow W_{j} - \gamma \, \frac{\partial {\mathcal{L}} }{\partial{W_{j}}} \label{eq:gradient_descent_2}
\end{equation}
\begin{equation}
    b_{j+1} \leftarrow b_{j} - \gamma \, \frac{\partial {\mathcal{L}} }{\partial{b_{j}}}  \label{eq:gradient_descent_3}
\end{equation}
where \(\gamma\) is the learning rate. Moreover, Adam optimization method \cite{kingma2014adam} which is a stochastic gradient descent method, BFGS \cite{zhao2021broyden} which is a quasi-Newton optimization algorithm, or L-BFGS (Limited-memory BFGS) optimizers \cite{xiao2008limited} which is an extension of BFGS, can be used.

\subsection{Energy Approach - Deep Energy Method}  \label{sec:DEM}
An alternative method for addressing the beam problems involves formulating it as a variational problem by applying the principle of minimum total potential energy. It is derived as a specific instance of the principle of virtual displacement, provided that the constitutive relations can be derived from a potential function. In this context, we focus our analysis on materials that permit the presence of a strain energy potential, allowing the stress to be derived from it. These materials are commonly referred to as hyperelastic. 

The principle of minimum total potential energy is a statement of the fact that the energy of the system is the minimum only at its equilibrium configuration. On the other hand, given the fact that the training process in machine learning can be regarded as a process of minimizing the loss function, it seems natural to regard the energy of the system as a very good candidate for this loss function.

The total potential energy is defined as follows:
\begin{equation}
\Pi = U+V_E
\end{equation}
where \(V_E\) represents the potential energy resulting from external loads, and \(U\) denotes the strain energy. Consider a body occupying the volume \(\Omega\), which experiences a body force \(\hat{\mathbf{f}}\) (measured per unit volume), a prescribed surface traction \(\hat{\mathbf{t}}\) (measured per unit area) on portion \(\Gamma_\sigma\), and specified displacement \(\hat{\mathbf{u}}\) on a portion \(\Gamma_u\) of the total surface \(\Gamma\) of \(\Omega\). The total potential energy for the given problem is expressed as:

\begin{equation}
\Pi(\mathbf{u}) = \int_\Omega \, \left[ \frac{1}{2} \, \mathbf{\sigma} : \mathbf{\varepsilon} - \hat{\mathbf{f}} \cdot \mathbf{u} \right] \, d\Omega - \int_{\Gamma_\sigma} \, \hat{\mathbf{t}} \cdot \mathbf{u} d\Gamma
\end{equation}

and considering the Eqs. \ref{eq:Green_Lagrange_strain_tensor}, \ref{eq:Constitutive} and \ref{eq:Lame_parameters}, the following is obtained:

\begin{equation}
\begin{aligned}
\Pi(\mathbf{u}) = &\int_\Omega \, \left[  
(\mu+\frac{\lambda}{2}) \left( \left( \frac{\partial \mathcal{N}_x}{\partial x}\right)^2 + \left( \frac{\partial \mathcal{N}_z}{\partial z} \right)^2\right) 
+ \frac{\mu}{4}\, \left( \frac{\partial \mathcal{N}_x}{\partial z} \right)^2   
+ \frac{\mu+\lambda}{4}\, \left( \frac{\partial \mathcal{N}_z}{\partial x} \right)^2 \right] d\Omega \\
+ & \int_\Omega \, \left[ \,
\lambda\, \frac{\partial \mathcal{N}_x}{\partial x}\, \frac{\partial \mathcal{N}_z}{\partial z} 
+ \frac{\mu}{2}\, \frac{\partial \mathcal{N}_x}{\partial z}\, \frac{\partial \mathcal{N}_z}{\partial x} \right] d\Omega \\
+ &\int_\Omega \, \left[
(\mu+\frac{3\lambda}{4}) \, \frac{\partial \mathcal{N}_x}{\partial x}\, \left(\frac{\partial \mathcal{N}_z}{\partial x}\right)^2 
+ \frac{\lambda}{4} \, \frac{\partial \mathcal{N}_z}{\partial z}\, \left(\frac{\partial \mathcal{N}_z}{\partial x}\right)^2 
+ \frac{1}{4}(\mu+\lambda) \, \left(\frac{\partial \mathcal{N}_z}{\partial x}\right)^4
\right] d\Omega \\
- &\int_\Omega \, \left(
\hat{f}_x \, \mathcal{N}_x + \hat{f}_z \, \mathcal{N}_z
\right) d\Omega - 
\int_{\Gamma_\sigma} \left(
\hat{t}_x u_x + \hat{t}_z u_z \, \right) d\Gamma
\label{eq:loss_energy}
\end{aligned}
\end{equation}

The above integral can be considered as the loss function and so the parameters of the networks can be extracted by the following:
\begin{equation}
    W_{j+1} \leftarrow W_{j} - \gamma \, \frac{\partial \Pi }{\partial{W_{j}}} \label{eq:gradient_descent_4}
\end{equation}
\begin{equation}
    b_{j+1} \leftarrow b_{j} - \gamma \, \frac{\partial \Pi }{\partial{b_{j}}}  \label{eq:gradient_descent_5}
\end{equation}

It is evident that the approach outlined possesses advantages when compared to the vector approach, because it exclusively requires the computation of first derivatives. On the other hand, the energy functional must be evaluated by numerical integration, which requires a quadrature method. This is generally not difficult for the geometries considered here. 

\subsection{Data-Driven Approach - Neural Operator} \label{sec:NOs}
Neural Networks, which are discussed in the sections \ref{sec:PINNs} and \ref{sec:DEM}, in fact, have been designed to learn mappings between coordinates of points in the domain into the displacement vector for a specific problem, with determined boundary conditions. The concept of Neural Operators represents a significant extension of traditional neural network development, by learning operators, that map between infinite-dimensional function spaces. Hence, Neural Operators can address a range of problems instead of being limited to a singular one, offering a computationally efficient solution. For instance, consider rephrasing Eq. \ref{eq:Piola-Kirchhoff} in the following manner:
\begin{equation}
L(a, \mathbf{u}) \coloneqq - \mathbf{\nabla} \cdot \left[ \mathbf{S} \cdot \left( \mathbf{I} + \mathbf{\nabla u} \right) \right] = \hat{\mathbf{f}} \label{eq:NeuralOperator}
\end{equation}

where \(a \) is a specific parameter from the set  \(\mathcal{A}\), \(\hat{\mathbf{f}} \) is any element from the dual space \(\mathcal{U}^*\) and \(D \) belongs to \(\mathbb{R}^d\). We assume that the solution \(\mathbf{u} : D\rightarrow \mathbb{R} \) belongs to the Banach space \(\mathcal{U}\) and \(L(a, \mathbf{u}) \rightarrow \hat{\mathbf{f}}\) is a forward mapping between the solution \(\mathbf{u}\), and problem parameters \(a\), and the problem data \(\hat{\mathbf{f}}\). The parameter \(a\), could indicate, e.g. the distribution of the material density over the problem domain. An operator that can be derived from this PDE is \( \mathcal{G}^{\dagger} \coloneqq L^{-1}( a, \hat{\mathbf{f}} ) : \mathcal{A} \times \mathcal{U} \rightarrow \mathcal{U}\) defined to map the body force to the corresponding solution \( \hat{\mathbf{f}} \mapsto \mathbf{u}\). 

Our goal is to learn a mapping between two infinite dimensional spaces by using a finite collection of observations of input-output pairs from this mapping. Given a collection of paired solutions \( \left\{ \left(a^{(i)}, \hat{\mathbf{f}}^{(i)}\right), \mathbf{u}^{(i)} \right\}\), where \( \mathbf{u}^{(i)} = \mathcal{G}^{\dagger} \left(a^{(i)},  \hat{\mathbf{f}}^{(i)} \right) \), our goal is to construct an approximation of \( \mathcal{G}^{\dagger} \) through the parametric map

\begin{equation}
\mathcal{G}_{\theta} : \mathcal{A} \times \mathcal{U} \rightarrow \mathcal{U}, \quad \theta \in \mathbb{R}^p
\end{equation}
with parameter from the finite-dimensional space \(\mathbb{R}^p\) and the choosing \(\theta^{\dagger} \in \mathbb{R}^p\) so that \(\mathcal{G}_{\theta^{\dagger}} \approx \mathcal{G}^{\dagger}\). 

We are focused on managing the average error of the approximation, specifically, our objective is to minimize the following norm of the approximation
\begin{equation}
\lVert \mathcal{G}^{\dagger}-\mathcal{G}_{\theta} \rVert^2_{L^2} 
 = \min_{\theta \, \in \, \mathbb{R}^p} \frac{1}{N} \sum_{i=1}^{N} \, \lVert \mathbf{u^{(i)}}-\mathcal{G}_{\theta}(\hat{\mathbf{f}}^{(i)}) \rVert^2_{\mathcal{U}}
\end{equation}

Approximating the \( \mathcal{G}^{\dagger} \) operator poses a distinct and generally more difficult challenge compared to determining the solution \( \mathbf{u} \in \mathcal{U} \) for a singular occurrence of the parameter \( a \in \mathcal{A}\) or \( \hat{\mathbf{f}} \in \mathcal{U}\). Most existing approaches, such as classical finite elements, finite differences, and finite volumes, as well as PINNs (discussed in section \ref{sec:PINNs}) and DEM (discussed in section \ref{sec:DEM}), are geared towards the latter task and consequently tend to be computationally inefficient. This causes them to be suboptimal for scenarios where the problem's solution is needed across numerous instances of the parameter. Conversely, the Neural Operator directly estimates the operator, resulting in a significantly more economical and quicker process, once the model is trained, leading to substantial computational efficiency gains compared to conventional solvers. There are some Neural Operator architectures, that can be used, such as Multi-Wavelet neural operator \cite{gupta2021multiwavelet},  U-shaped Neural Operator \cite{ashiqur2022u}, Multipole Graph Neural Operator (MGNO) \cite{li2020multipole}, Fourier Neural Operator (FNO) \cite{li2021fourier}, and DeepONet \cite{Lu2021DeepONet}.

Therefore, if we consider the displacement field of the beam as follows:
\begin{equation}
\begin{aligned}
& \mathbf{u}_x(x, z) = \mathcal{G}_{\theta_x} (\hat{\mathbf{f}};\theta)(x,z) \\
& \mathbf{u}_y(x, z) = 0 \\
& \mathbf{u}_z(x, z) = \mathcal{G}_{\theta_z} (\hat{\mathbf{f}};\theta)(x,z)
\label{eq:displacement_field_NO}
\end{aligned}
\end{equation}
all of the Eqs. \ref{eq:nonzero_strain_tensor}, \ref{eq:rotation_vector}-\ref{eq:sigm_xz} and \ref{eq:M_xx}-\ref{eq:gradient_descent_3} can be rewritten by substitution of \( \mathcal{G}_{\theta_x} \) and \( \mathcal{G}_{\theta_z} \) for \(\mathcal{N}_x\) and \(\mathcal{N}_y\) respectively. 

In conclusion, we have considered three primary methodologies. The efficacy of a particular approach within DNB depends on the specific characteristics of the problem at hand.

\section{Numerical Results} \label{sec:Result}
In this section, we explore the application of DNB to solve some different problems in FG porous beam analysis and present their result. The first example examines an end-loaded cantilever, a commonly employed case in the literature, and compares the results obtained using DNB with the available Timoshenko exact solution. In the next example, DNB is implemented to analyze an FG porous beam, and later, we consider the application of DNB in training a Neural Operator to predict an FG porous beam with any material distribution and any arbitrary traction. The code and accompanying data will be publicly available \href{https://github.com/eshaghi-ms/DeepNetBeam}{https://github.com/eshaghi-ms/DeepNetBeam}. Specifics regarding the network architecture, including the number of layers, neurons in each layer, and the chosen activation functions, have been provided for each example. The computational times have been obtained by training the neural network on an NVIDIA A100-PCIE-40GB GPU.

\subsection{Cantilever Beam}
\begin{figure}
    \centering
    \begin{circuitikz}
    \fill[pattern=north east lines] (0,0.75) rectangle (0.25,3.25);
    \draw (0.25,0.75) -- (0.25,3.2);
    \draw[fill=gray!40] (0.25,1) rectangle (8,3);
    \draw[<->|] (0.25,0.4) -- (8,0.4) node[midway,below] {$L$};
    \draw[-{Straight Barb[left,length=3mm,width=2mm]}] (8.2,3.0)  -- (8.2,1) node[midway,right] {$p(x, z)$};
    \draw[|<->|] (9.7,1) -- (9.7,3) node[midway,right] {$D$};
    \end{circuitikz}
    \caption{Cantilever Beam Example}
    \label{fig:Cantilever beam example}
\end{figure}
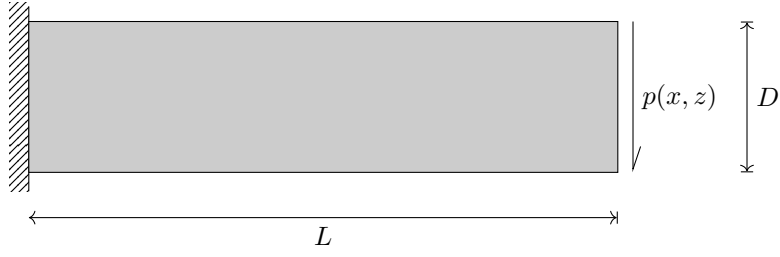

In this section, to show the implementation of DNB in vector approach, which is discussed in section \ref{sec:PINNs}, we consider a simple example of a cantilever beam, in Fig. \ref{fig:Cantilever beam example} with dimensions including depth \(D\), length \(L\), and unit thickness. This beam has prescribed displacements at \(x = 0\) and sustains an end load \(p(z)\). Therefore, the equilibrium equation is 
\begin{equation}
-\nabla \cdot \sigma(x,z) = f(x,z) \quad \text{for } x, z \in \Omega 
\end{equation}
with the strain-displacement equation:
\begin{equation}
    \varepsilon(x,z) = \frac{1}{2}(\nabla u + \nabla u^T)
\end{equation}
and the constitutive law in Eq. \ref{eq:Constitutive}, where \(\sigma\) is the stress tensor, \(\varepsilon\) is strain tensor, \(f\) is body force, \(u = (u_x, u_z)\) is displacement field, \(\Omega\) is problem domain, and with the Dirichlet boundary conditions: 
\begin{equation}
u(x,z)=\hat{u}(x,z) \quad \text{for } x, z\in\Gamma_D
\end{equation}
where \(\hat{u}(x,z)\) is the displacement at the boundary and Neumann boundary conditions: 
\begin{equation}
\sigma \cdot n = \hat{t}\,(x,z) \quad \text{for } x,z\in \Gamma_N,
\end{equation}
where \(\hat{t}(x,z)\) is traction at the boundary and \(n\) is the normal vector.

We solve the problem when \(\Omega\) is a rectangle with corners at  (0,0) and (8,2) \si{\meter}, Dirichlet boundary conditions for \(x=0\):
\begin{equation}
u_x(z) = \frac{Pz}{6EI_{yy}}  \left((2+\nu)(z^2-\frac{D^2}{4})\right)
\end{equation}
\begin{equation}
u_z(z) = -\frac{\nu}{2}\frac{PL z^2}{EI_{yy}}
\end{equation}
and parabolic traction at \(x=L\)
\begin{equation}
\hat{t}(z) = P \left( \frac{z^2 - D z}{2I_{yy}} \right)
\end{equation}
where \(P=\SI{2}{\mega\newton}\) is the maxmimum traction, \(E = 10^5 \si{\mega\pascal}\) is Young's modulus, \(\nu = 0.25\) is the Poisson ratio, and \(I_{yy} = D^3/12\) is second moment of area of the cross-section.
For the validation of results, Timoshenko and Goodier \cite{timoshenko1951theory} have demonstrated that the stress distribution within the cantilever can be described as follows:

\begin{equation}
\sigma_{xx} = \frac{P (L-x) z}{I_{yy}}
\end{equation}
\begin{equation}
\sigma_{zz} = 0
\end{equation}
\begin{equation}
\tau_{xz} = - \frac{P}{2I_{yy}} \left( \frac{D^2}{4} - z^2\right)
\end{equation}

and the displacement field \( \{u_x,u_z\} \) is given by:
\begin{equation}
u_x = -\frac{Pz}{6EI_{yy}} \left(  ( 6L - 3x ) x + ( 2 + \nu) \left( z^2 - \frac{D^2}{4} \right) \right)
\end{equation}
\begin{equation}
u_z = -\frac{P}{6EI_{yy}} \left(  3 \nu z^2 ( L - x ) + ( 4 + 5\nu) \frac{D^2x}{4} + (3L - x)x^2 \right)
\end{equation}. 

\begin{figure}[htbp]
    \centering
    \includegraphics[width=1\textwidth]{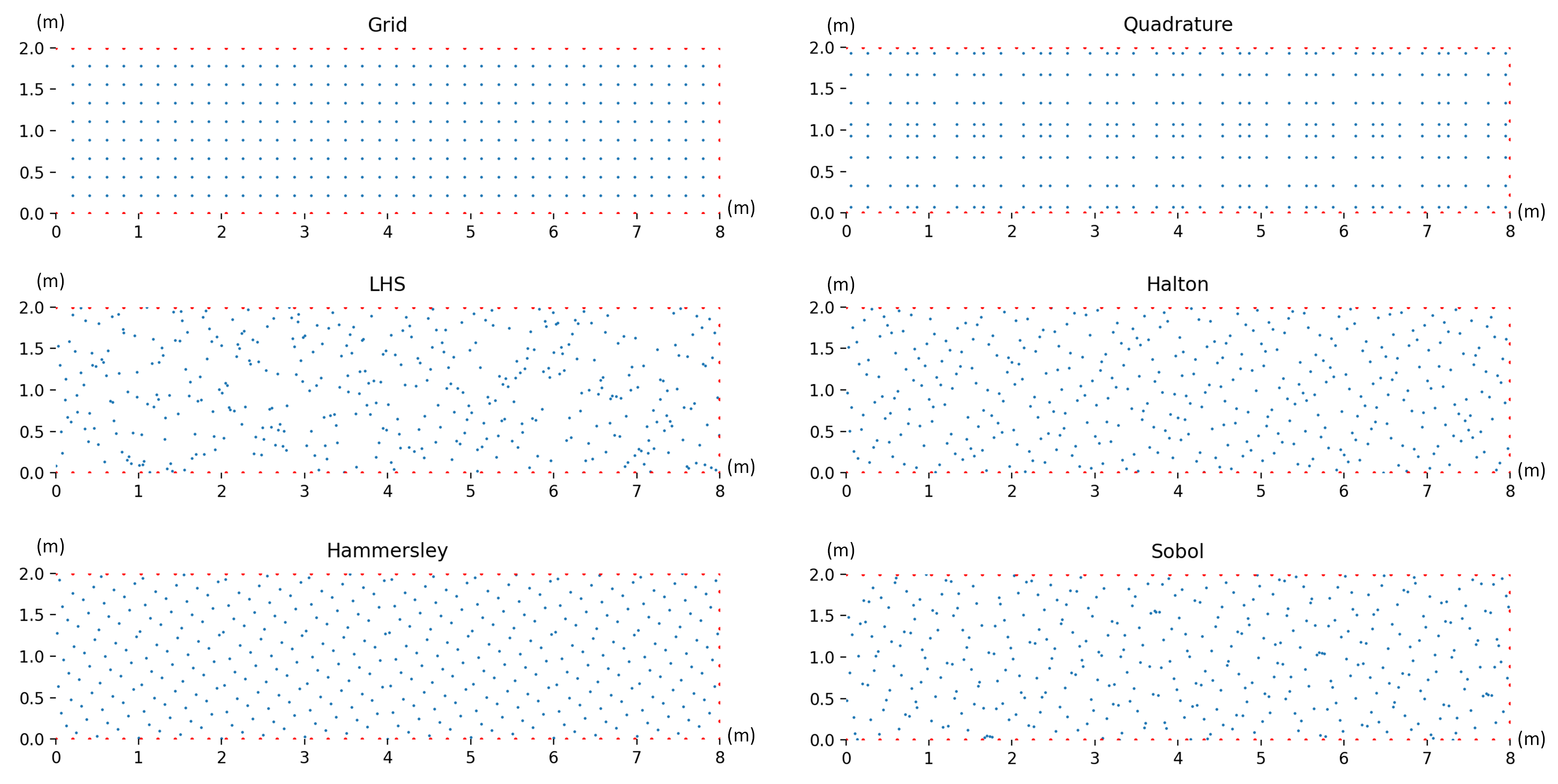}
    \vspace{-10pt}
    \caption{Examples of 400 points generated in \([0, 2]\times [0, 8]\) using different sampling methods}
    \label{fig:sampling_methods}
\end{figure}

\begin{figure}[htbp]
    \centering
    \includegraphics[width=0.8\textwidth]{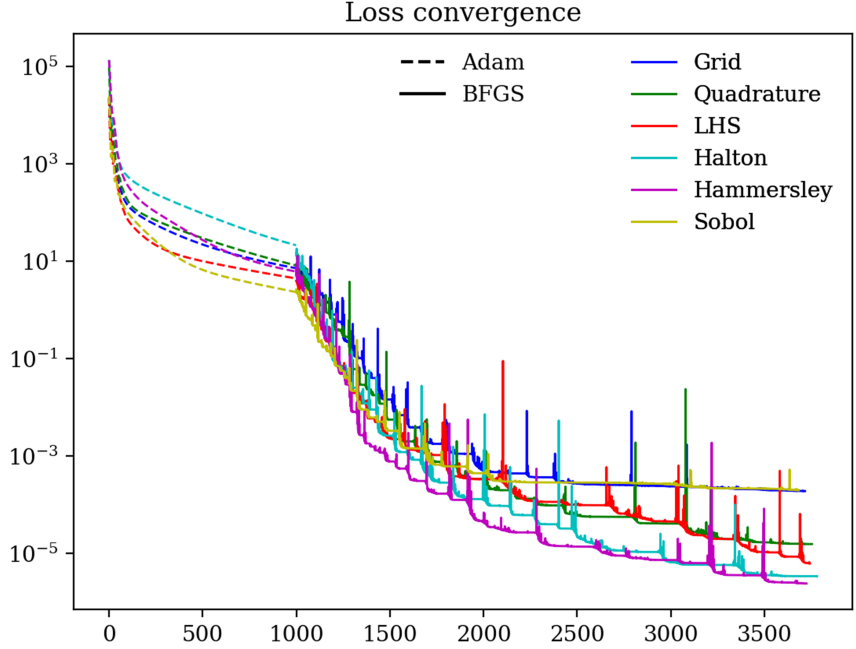}
    \vspace{-10pt}
    \caption{Convergence of the loss function in different sampling methods}
    \label{fig:loss_converge_Cantilever}
\end{figure}

To obtain the displacement field, we have used a network with 3 hidden layers of 20 neurons each and considered the swish activation function across all layers \cite{ramachandran2017searching}. For sampling strategy \cite{wu2023comprehensive} of choosing the collocation points for the enforcement of physical constraints, we have investigated six methods as follows and the examples of 400 points generated in \([0,2]\times[0,8]\) using the methods are shown in Fig. \ref{fig:sampling_methods}: 

\begin{itemize}
\item 1. \textbf{Equispaced Uniform Grid (Grid)}: Residual points are chosen at evenly spaced intervals across the computational domain, forming a uniform grid.
\item 2. \textbf{Gaussian Quadrature Points (Quadrature)}: Residual points are selected based on the famous \(n\)-point Gaussian quadrature rule, which is constructed to yield an exact result for the integral of polynomials of degree \(2n-1\) or less by a suitable choice of the nodes (in the current work \(n = 4\)).
\item 3. \textbf{Latin Hypercube Sampling (LHS)}\cite{mckay2000comparison}: A Monte Carlo method that generates random samples within intervals based on equal probability, ensuring that the samples are normally distributed within each range.
\item 4. \textbf{Halton Sequence (Halton)}\cite{halton1960efficiency}: Samples are generated by reversing or flipping the base conversion of numbers using prime bases.
\item 5. \textbf{Hammersley Sequence (Hammersley)}\cite{hammersley1964monte}: Similar to the Halton sequence, but with points in the first dimension spaced equidistantly.
\item \ 6. \textbf{Sobol Sequence (Sobol)}\cite{sobol1967distribution}: A base-2 digital sequence that distributes points in a highly uniform manner.
\end{itemize}

\begin{table}[ht]
\centering
\caption{$L^2$ relative error for different sampling methods}
\vspace{-10pt}
\label{relative_error}
\begin{tabular}{>{\raggedright}p{2.5cm}p{2.5cm}c}
\hline
 & Train data & Test data \\
\hline
Grid       & 0.472 \% & 0.465 $\pm$ 0.210 \% \\
Quadrature & 0.220 \% & 0.218 $\pm$ 0.052 \% \\
LHS        & 0.049 \% & 0.048 $\pm$ 0.016 \% \\
Halton     & 0.011 \% & 0.012 $\pm$ 0.007 \% \\
Hammersley & 0.024 \% & 0.024 $\pm$ 0.010 \% \\
Sobol      & 0.174 \% & 0.178 $\pm$ 0.104 \% \\
\hline
\end{tabular}
\end{table}

\begin{figure}[htbp]
    \centering
    \makebox[\textwidth]{\includegraphics[width=1.3\textwidth]{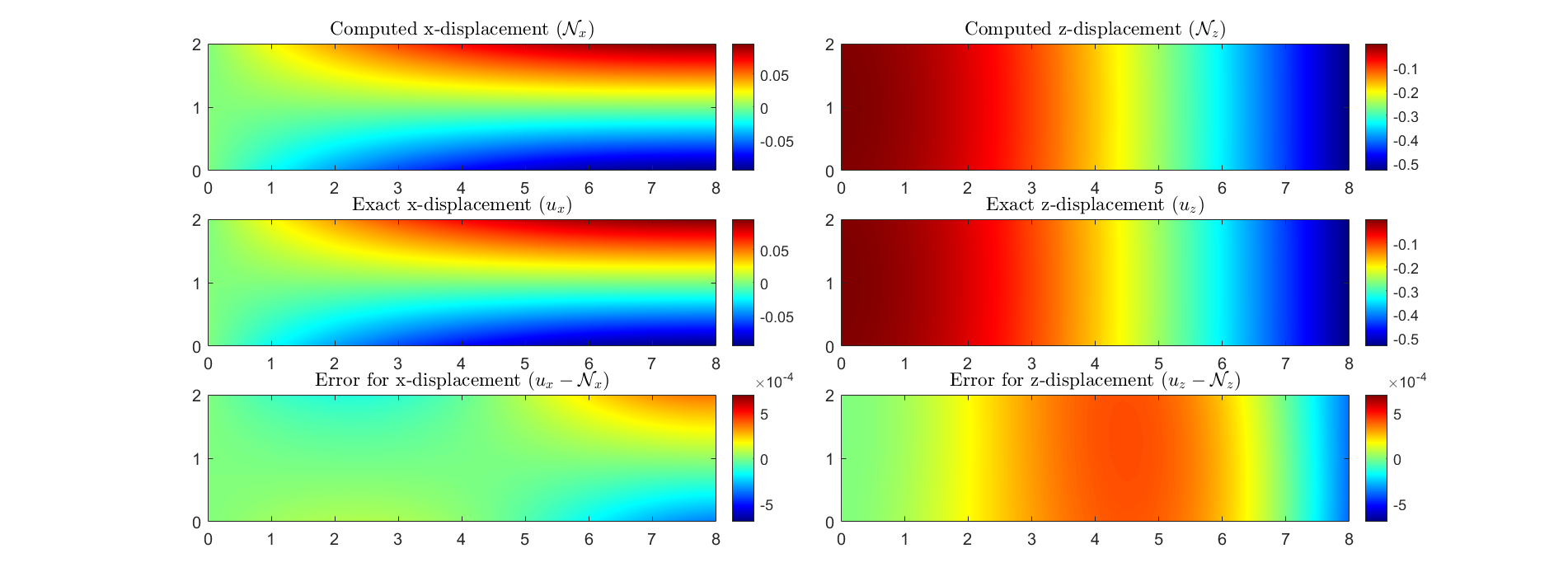}}
    \vspace{-40pt}
    \caption{Computed solution for the Cantilever beam \((\si{cm})\).}
    \label{fig:Result_Cantilever}
\end{figure}

For optimization, the neural network is trained by using a combination of Adam optimizer and second-order quasi-Newton method (BFGS) and the relative error in \(\mathcal{L}_2\) norm for the displacement, which is calculated as 

\begin{equation}
    \mathcal{L}_2^{\text{rel}} = \frac{\sqrt{\sum_{i=1}^{N_d} \left[ \left( \mathcal{N}_x - \hat{u}_x \right)^2 + \left( \mathcal{N}_z - \hat{u}_z \right)^2 \right] }}{\sqrt{\sum_{i=1}^{N_d} \left( \hat{u}_x^2 + \hat{u}_z^2 \right) }}
    \label{eq:L2_error}
\end{equation}

is equal to \(0.47\%\), \(0.22\%\), \(0.05\%\), \(0.01\%\), \(0.02\%\) and \(0.17\%\) respectively for Grid, Quadrature, LHS, Halton, Hammersley, and Sobol sampling method with a computational time of 96 s. Among the six sampling methods, Halton and Hammersley, which are low-discrepancy sequences, generally perform better than the others, as shown in Table \ref{relative_error}. Although the Grid method yields the lowest accuracy, all methods produce sufficiently low errors due to the smoothness of the problem's PDE. Fig. \ref{fig:loss_converge_Cantilever} shows the convergence of the loss function, and the final results for Quadrature sampling method are demonstrated in Fig. \ref{fig:Result_Cantilever}.

\subsection{Functionally Graded Porous Beam}

This example presents the bending analysis of functionally graded porous beams to implement the energy approach in DNB, which is discussed in section \ref{sec:DEM}. The example assumes that porous composites' elasticity moduli and mass density vary in thickness based on two distinct distribution patterns. The mechanical properties of an open-cell metal foam are examined as a representative case to establish the correlation between density and porosity coefficients and the porous beams' bending behavior is described by a system of PDEs. The example explores three different boundary conditions, including a beam with hinged-hinged (H-H), clamped-clamped (C-C), and clamped-hinged (C-H) end supports. 

\begin{figure}[htbp]
    \centering
    \includegraphics[width=1\textwidth]{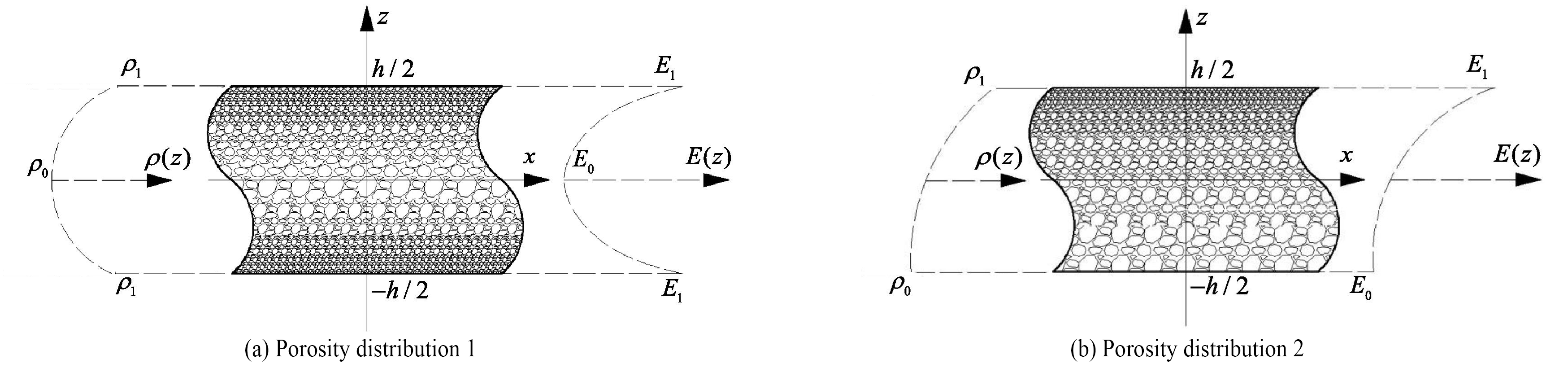}
    \caption{Two porosity distribution patterns.}
    \label{fig:porosity_distribution}
    
    \vspace{10pt} 
    
    \raggedright 
    Porosity distribution 1 (Symmetric): 
    \begin{equation}
        E(z) = E_1 \left[ 1-e_0 \cos(\pi \zeta) \right], \quad
        G(z) = G_1 \left[ 1-e_0 \cos(\pi \zeta) \right], \quad
        \rho(z) = \rho_1 \left[ 1-e_m \cos(\pi \zeta) \right]
    \label{eq:porosity_distribution_1} 
    \end{equation}
    
    Porosity distribution 2 (Asymmetric): 
    \begin{equation}
        E(z) = E_1 \left[ 1-e_0 \cos\left(\frac{\pi}{2}\zeta +\frac{\pi}{4}\right) \right], \quad  
        G(z) = G_1 \left[ 1-e_0 \cos\left(\frac{\pi}{2}\zeta +\frac{\pi}{4}\right) \right], \quad  
        \rho(z) = \rho_1 \left[ 1-e_m \cos\left(\frac{\pi}{2}\zeta +\frac{\pi}{4}\right) \right]
    \label{eq:porosity_distribution_2}
    \end{equation}
\end{figure}

\begin{figure}[htbp]
    \centering
    \includegraphics[width=0.9\textwidth]{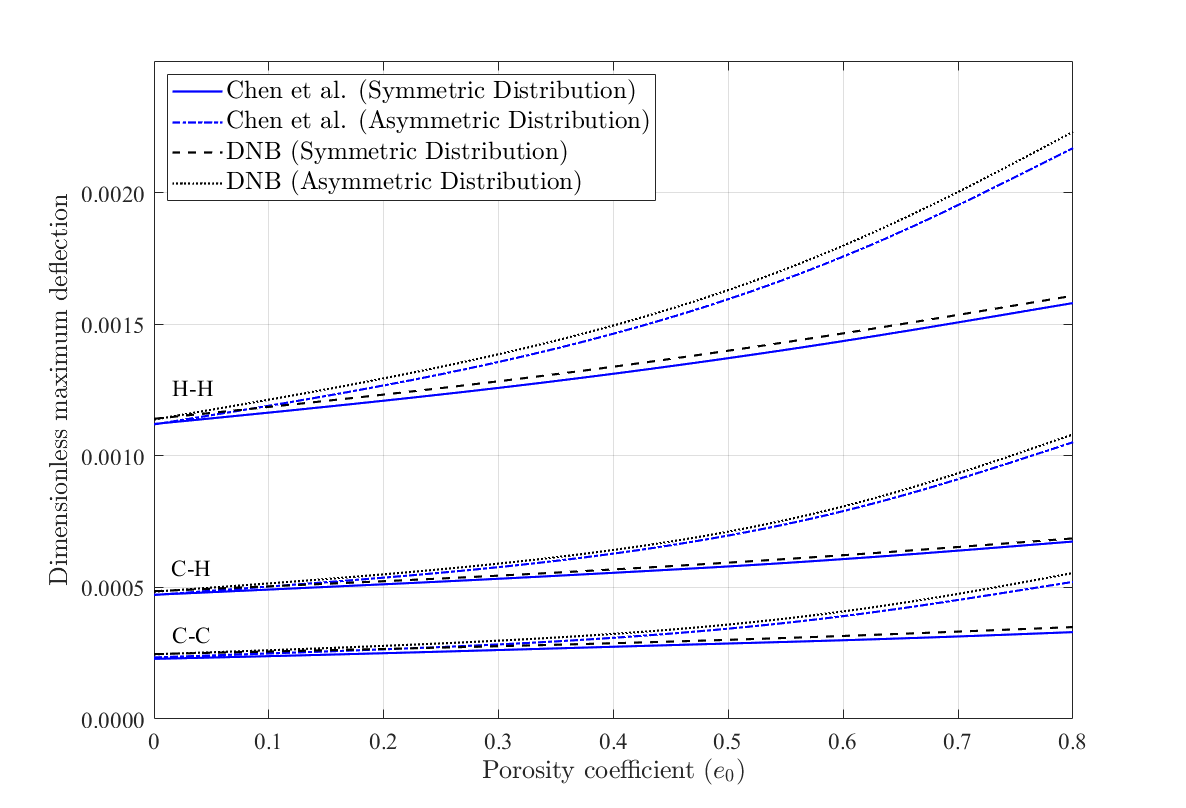}
    \vspace{-10pt}
    \caption{Dimensionless maximum deflection under a uniformly distributed load \((L/h = 20)\).}
    \label{fig:Validation_Deflection}
\end{figure}

An FG porous beam of thickness \(h\) and length \(L\) with two different porosity distributions along the thickness direction is shown in Fig. \ref{fig:porosity_distribution}.a for porosity distribution 1 \cite{magnucki2004elastic} and Fig. \ref{fig:porosity_distribution}.b for porosity distribution 2 \cite{jabbari2014buckling}. The beam is represented in a rectangular coordinate system, with the \(z\)-axis indicating the thickness direction and the \(x\)-axis indicating the length direction. Owing to the non-uniform porosity distribution, Young's modulus, shear modulus, and mass density exhibit smooth variations, as described by Eqs. \ref{eq:porosity_distribution_1} and Eqs. \ref{eq:porosity_distribution_2} for porosity distributions 1 and 2, respectively. Both distributions share the same maximum and minimum values for elasticity moduli and mass density. In distribution 1, the minimum values occur on the midplane of the beam, featuring the largest size and density of internal pores, while the maximum values are found on the top and bottom surfaces, equivalent to those of homogeneous beams made of pure materials. In distribution 2, elasticity moduli and mass density reach their maximum on the top surface and generally decrease towards the minimum values on the bottom surface.

\begin{figure}[htbp]
    \centering
    \makebox[\textwidth]{\includegraphics[width=1.3\textwidth]{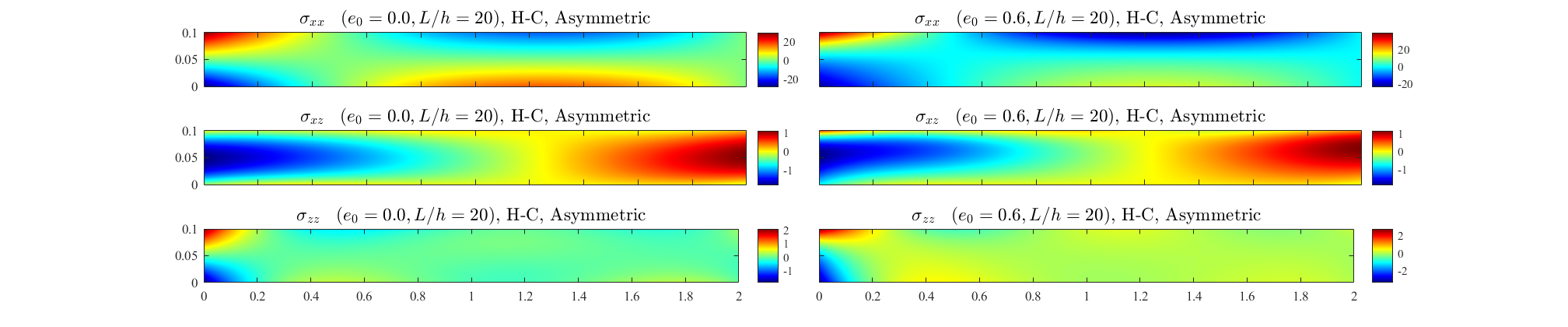}}
    \vspace{-30pt}
    \caption{Stress \((\si{\mega\pascal})\) components distribution for non-porous and porous (asymmetric) beam \((L/h = 20)\). }
    \label{fig:Stress_components_distribution}
\end{figure}

Here, \(\zeta = z/h\), the porosity coefficient, \(e_0\) is defined as \(e_0=1-E_0/E_1=1-G_0/G_1 \, \) with \(0<e_0<1\). The minimum and maximum values of Young's modulus \(E_0\) and \(E_1\) are related to the minimum and maximum values of shear modulus \(G_0\) and \(G_1\) by \(G_i=E_i/(2(1+\nu))\) where \(\nu\) is the Poisson's ratio, a constant across the beam thickness. The porosity coefficient for mass density is defined as \(e_m = 1-\rho_0/\rho_1 \, (0<e_m<1)\) in which \(\rho_0\) and \(\rho_1\) represent the minimum and maximum values of mas density, respectively. Therefore, the loss function for the porous beam is obtained by putting the equation \ref{eq:Lame_parameters} into the Eq. \ref{eq:loss_energy}.

To obtain the displacement field, a neural network with three hidden layers, each consisting of 20 neurons, was employed. The network was trained on a grid of uniformly spaced points \(N_x \times N_z\) within the domain's interior, where \(N_x = 80\) and \(N_z = 40\). The swish activation function was applied to all layers. Optimization was performed through a combination of the Adam optimizer and the second-order quasi-Newton method (BFGS) and the computational time taken is 22 s.

\begin{figure}[htbp]
    \centering
    \makebox[\textwidth]{\includegraphics[width=1.3\textwidth]{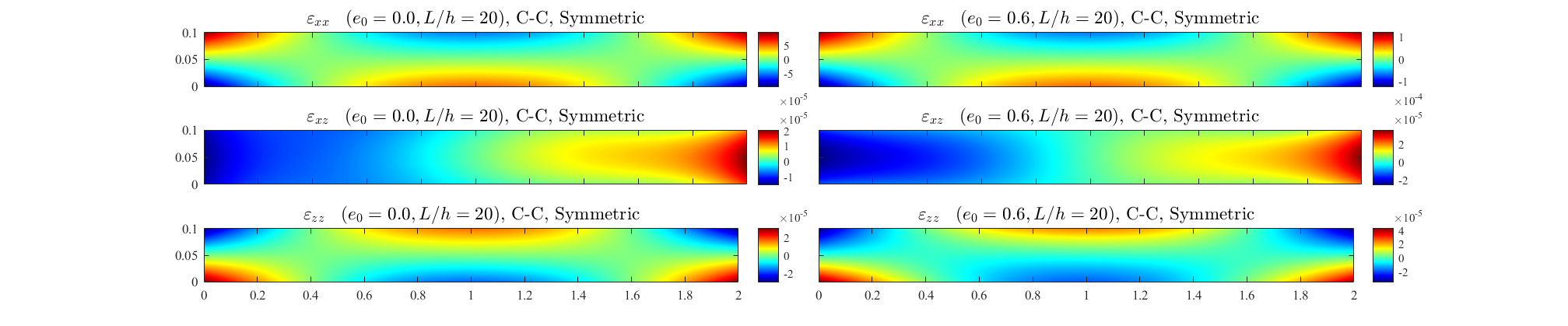}}
    \vspace{-30pt}
    \caption{Strain components distribution for non-porous and porous (symmetric) beam \((L/h = 20)\). }
    \label{fig:Strain_components_distribution}
\end{figure}

The material assumed for the porous beam is steel foam, characterized by  \(E_1 = \SI{200}{\giga\pascal}\), \(\nu = 1/3\), \(\rho_1 = \SI{7850}{\kilogram\per\meter\cubed}\). The beam's cross-section has dimensions  \(h = \SI{0.1}{\meter}\) and \(b = \SI{0.1}{\meter}\) (where \(b\) represents the width of the beam). The validation analysis is done through a direct comparison between the present results and the work by Chen et al. \cite{chen2015elastic}. Fig. \ref{fig:Validation_Deflection} shows the dimensionless bending deflections \(\Bar{u}_y\ = u_y / h\) under a distributed load \(Q = \SI{1e4}{\newton\per\meter}\) at the top of the beam and slenderness ratio \(L/h=20\) with varying porosity coefficients. Our results are in good agreement to those reported in Chen's study.

\begin{figure}[htbp]
    \centering
    \makebox[\textwidth]{\includegraphics[width=1.3\textwidth]{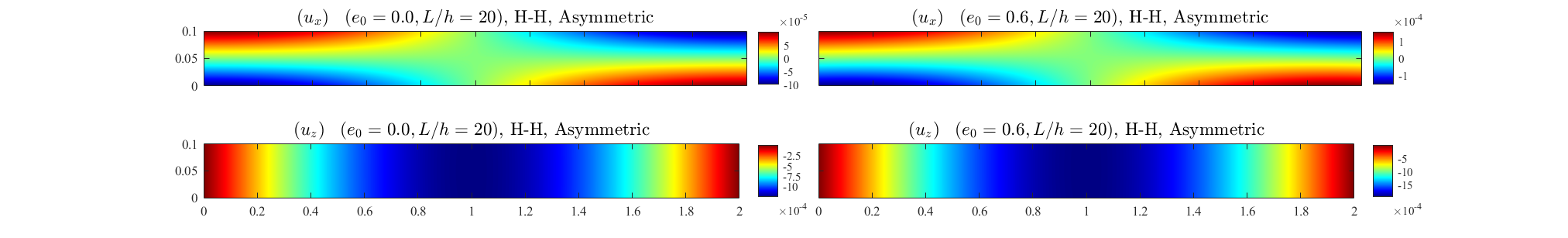}}
    \vspace{-30pt}
    \caption{Displacement \((\si{\meter})\) field distribution for non-porous and porous (asymmetric) beam \((L/h = 20)\) }
    \label{fig:Displacement_field}
\end{figure}

To observe variations in the components of stress, strain, and displacement fields within the beam, examples are provided in Figures \ref{fig:Stress_components_distribution}, \ref{fig:Strain_components_distribution}, and \ref{fig:Displacement_field}. Fig. \ref{fig:Stress_components_distribution} displays the distribution of stress components for an asymmetric beam with an H-C boundary condition. Fig. \ref{fig:Strain_components_distribution} illustrates the distribution of strain components for a symmetric beam with a C-C boundary condition. Finally, Fig. \ref{fig:Displacement_field} depicts the distribution of the displacement field for an asymmetric beam with an H-H boundary condition. It should be noted that a new neural network, with the same hyperparameters, is trained for each boundary condition.

\subsection{FG Beams with arbitrary traction and porosity distribution}
In this portion, we illustrate the Neural Operator in the data-driven approach discussed in section \ref{sec:NOs}. To exemplify this, Fourier Neural Operator (FNO) has been trained to learn an operator, that maps from arbitrary traction and porosity distribution functions into the displacement field of a beam with depth \(D\) and length \(L\). In other words, the networks are employed to solve a range of problems with any different arbitrary traction functions and porosity distribution, which has been shown in Fig. \ref{fig:beam_random_load}. The beam is considered to be attached to supports at \(x=0\) and \(x=L\) and sustains a distributed load on the top edge of the beam. As a result, Eq. \ref{eq:NeuralOperator} becomes as follows:

\begin{figure}
\centering
\begin{circuitikz}
    \fill[pattern=north east lines] (0,0.25) rectangle (0.25,2.25);
    \draw (0.25,0.25) -- (0.25,2.25);
    \fill[pattern=north east lines] (8.25,0.25) rectangle (8.5,2.25);
    \draw (8.25,0.25) -- (8.25,2.25);
    
    \draw[fill=gray!0] (0.25,1) plot [smooth, tension=1] coordinates {(0.25,3.07)  (1.25,2.95) (2.25,3.7) (3.25,3.07) (4.25,3.07) (5.25,3.4) (6.25,3.00) (7.25,3.45) (8.25,3.2)};

    \draw[pattern=grid] (0.25,0.5) rectangle (8.25,2);

    \draw[<->|] (0.25,0.0) -- (8.25,0.0) node[midway,below] {$L$};
    \draw[|<->|] (9,0.5) -- (9,2) node[midway,right] {$D$};
    
    \pgfmathsetseed{1} 
    \foreach \i in {0.25,1.25,...,7.25, 8.25}
    {
        \pgfmathsetmacro{\randomload}{-abs(rand*1)-0.7} 
        \draw[latex-, dashed, line width=0.7pt] (\i,2.1) -- ++(0,-\randomload) node[pos=0.5,right] {};
    }
    \node[align=center] at (4.125, 3.7) {\footnotesize{Arbitrary Traction}};

    \draw[fill=gray!0] (9.25,1) plot [smooth, tension=1] coordinates {(11.14,0.5)  (11.54,1.00) (10.85,1.5) (11.3,2.0)};

    \pgfmathsetseed{2} 
    \foreach \i in {0.5,1,...,2}
    {
        \pgfmathsetmacro{\randomE}{-abs(rand*1)-0.7} 
        \draw[-latex, dashed, line width=0.7pt] (10,\i) -- ++(-\randomE,0) node[pos=0.5,right] {};
    }

    \node at (12.5, 1.6) {\footnotesize{Arbitrary}};
    \node at (12.5, 1.3) {\footnotesize{Porosity}};
    \node at (12.5, 1.0) {\footnotesize{Distribution}};
    
\end{circuitikz}
\caption{Beam with Random Distributed Load and Random Porosity Distribution}
\label{fig:beam_random_load}
\end{figure}
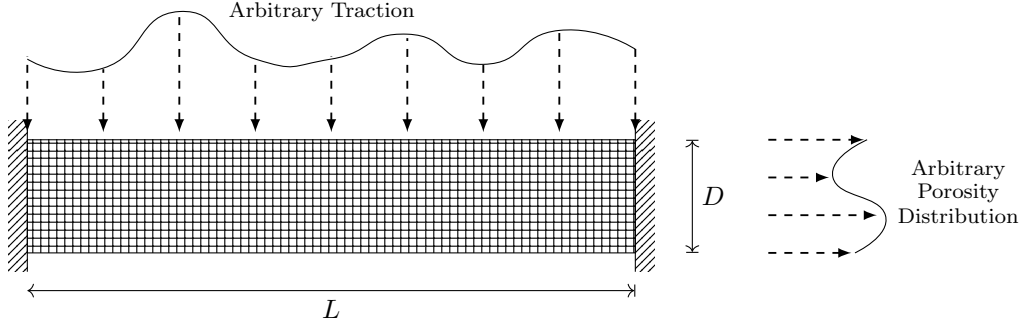

\begin{equation}
X \coloneqq \left\{ \begin{array}{ll}
- \mathbf{\nabla} \cdot \left[ \mathbf{S} \cdot \left( \mathbf{I} + \mathbf{\nabla u} \right) \right] = 0  \:\quad \text{for } x,\, z \in \Omega \\
\mathbf{u}(x,z) = 0 \,\qquad\qquad\qquad \text{ for } x=0, \, L\\
\sigma \cdot n = \hat{\mathbf{t}}\,(x) \qquad\qquad\qquad  \text{for } z=D\\
\mathbf{S} = \mathbf{S}(\hat{\mathbf{E}}(z)) \!\qquad\qquad\qquad  \text{for } x,\, z \in \Omega\\
\end{array} \right.
\end{equation}

The desired operator from this PDE is \(\mathcal{G}^{\dagger} \coloneqq L^{-1} \left\{\hat{\mathbf{t}}, \hat{\mathbf{E}}\right\} : \mathcal{U}^t \times \mathcal{U}^E \rightarrow \mathcal{U}\) defined to map the traction and porosity distribution to the corresponding displacement \( \left\{ \hat{\mathbf{t}}, \hat{\mathbf{E}} \right\} \mapsto \mathbf{u}\), \(\mathcal{U}^t\) is the space of continuous real-valued functions defined on the top boundary, \(\mathcal{U}^E\) is the space of continuous real-valued functions defined on \([0,L]*[0,D]\), and \(\mathcal{U}\) is the space of continuous functions with values in \(\mathbb{R}^2\), representing the \(x\)- and \(z\)-displacements. We solve the problem when the parameters, \(\Omega\) is a rectangle with corners at \((0,0)\) and \((2,0.1)\), \(E_1 = \SI{200}{\giga\pascal}\) is Young's modulus, \(\nu = 1/3\) is the Poisson ratio. For the data generation and validation, the isogeometric analysis (IGA) has been used \cite{anitescu2018recovery} and the result for various traction functions and different porosity distributions has been compared. To provide database including the pair solution \( \left\{ \hat{\mathbf{t}}^{(i)}, \hat{\mathbf{E}}^{(i)}, \mathbf{u}^{(i)} \right\}\), we have used a Gaussian Random Fields (GRF) \cite{williams2006gaussian} for creating random smooth functions with a length scale \(l=0.2\) for traction and a length scale \(l=0.025\) for porosity distribution as 
\begin{figure}[htbp]
    \centering
    \makebox[\textwidth]{\includegraphics[width=0.8\textwidth]{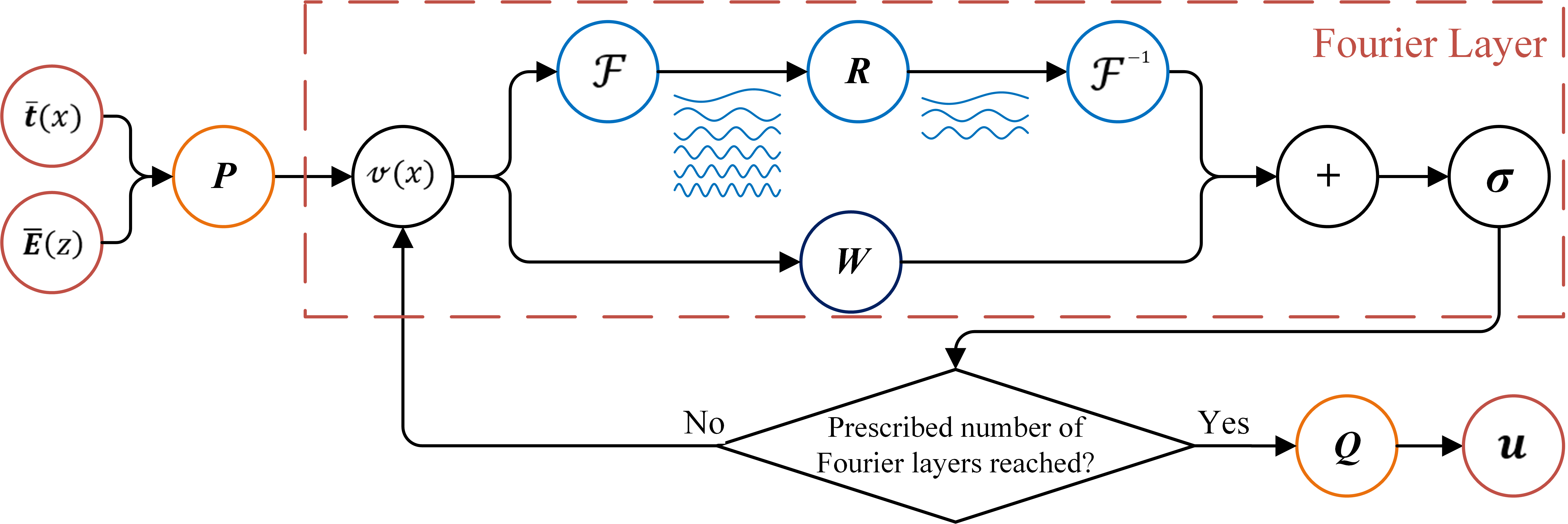}}
    \vspace{-20pt}
    \captionsetup{justification=centering,margin=0.5cm}
    \caption{The architecture of the Fourier neural operators}
    \label{fig:FNO}
\end{figure}

\begin{equation}
\begin{aligned}
& \hat{\mathbf{t}}(x) \sim \sigma_{t} \times \mathcal{GP}(0, k_1(x_1, x_2)) + \mu_{t} \\
& \hat{\mathbf{E}}(x) \sim \frac{E_{\text{max}}-E_{\text{min}}}{\mathcal{GP}_{\text{max}}-\mathcal{GP}_{\text{min}}} \times \left( \mathcal{GP}(0, k_1(x_1, x_2)) - \mathcal{GP}_{\text{min}} \right)+E_{\text{min}}
\label{eq:gaussian_random_fields}
\end{aligned}
\end{equation}
with an exponential quadratic covariance kernel \( k_1(x_1,x_2) = \text{exp}\left(-||x_1-x_2||^2/(2l^2)\right) \), where \(\sigma_{t}\) and \(\mu_t\) are expected variance and average of traction (here \(\sigma_{t} = 0.15\) and \(\mu_t = 0.2\)), \(E_{\text{min}}\) and \(E_{\text{max}}\) are expected values for minimum and maximum of elasticity modulus, which in the current example are selected randomly between \SI{20}{\giga\pascal} and \SI{380}{\giga\pascal}. Therefore our goal is to construct an approximation of \( \mathbf{u}^{(i)} = \mathcal{G}^{\dagger} \left( \hat{\mathbf{t}}^{(i)}, \hat{\mathbf{E}}^{(i)} \right) \) by the parametric map \(\mathcal{G}_{\theta} : \mathcal{U}^t \times \mathcal{U}^E \rightarrow \mathcal{U}, \quad \theta \in \mathbb{R}^p\). 

\begin{figure}[htbp]
    \centering
    \includegraphics[width=0.7\textwidth]{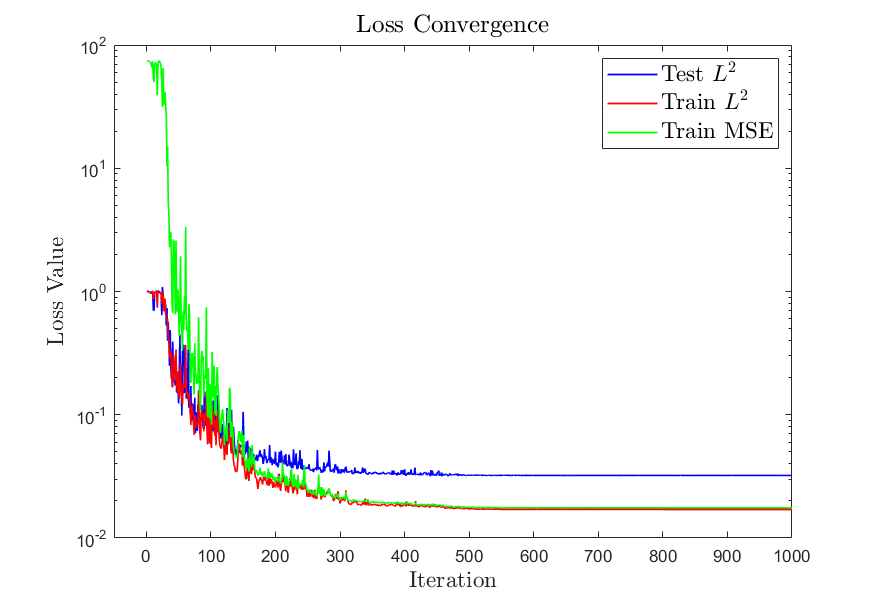}
    \vspace{-10pt}
    \caption{Convergence of the loss function.}
    \label{fig:loss_converge_FNO}
\end{figure}
To derive the parametric map \(\mathcal{G}_{\theta}\), we utilized an FNO structured as depicted in Fig. \ref{fig:FNO}. This FNO is constructed iteratively, denoted as \(v_0 \mapsto v_1 \mapsto \ldots \mapsto v_T\) where \(v_j\) for \(j=0,1, \ldots , T-1\) represents a sequence of functions (Convolutional neural network) with values in \(\mathbb{R}^{d_v}\). As illustrated in Fig. \ref{fig:FNO}, the inputs \(\hat{\mathbf{t}}\) and \(\hat{\mathbf{E}}\) undergo a lift to a higher-dimensional representation \(v_0(x,z) = P(\hat{\mathbf{t}})\) through a local transformation \(P\), which is implemented as a 2-layer fully-connected neural network with 32 neurons in each layer. Subsequently, four Fourier layers are used, transforming \(v_t\) to \(v_{t+1}\). The final output \(u(x,z) = Q(v_T(x,z))\) is obtained by projecting \(v_T\) through a local transformation \(Q:\mathbb{R}^{d_v} \mapsto \mathbb{R}^2\). In this case, \(Q\) is a 2-hidden-layer neural network with 128 in the first layer and 2 neurons in the second layer. In addition, the Fourier layer is characterized by the following equation:
\begin{equation}
v_{t+1}(\chi) \coloneqq \sigma \left( W\,v_t(\chi) + \left( \mathcal{K}(a;\phi)\,v_t \right) (\chi)\right), \quad \forall \chi \in \Omega \label{eq:FourierLayer}
\end{equation}
\begin{figure}[htb]
    \centering
    \makebox[\textwidth]{\includegraphics[width=1\textwidth]{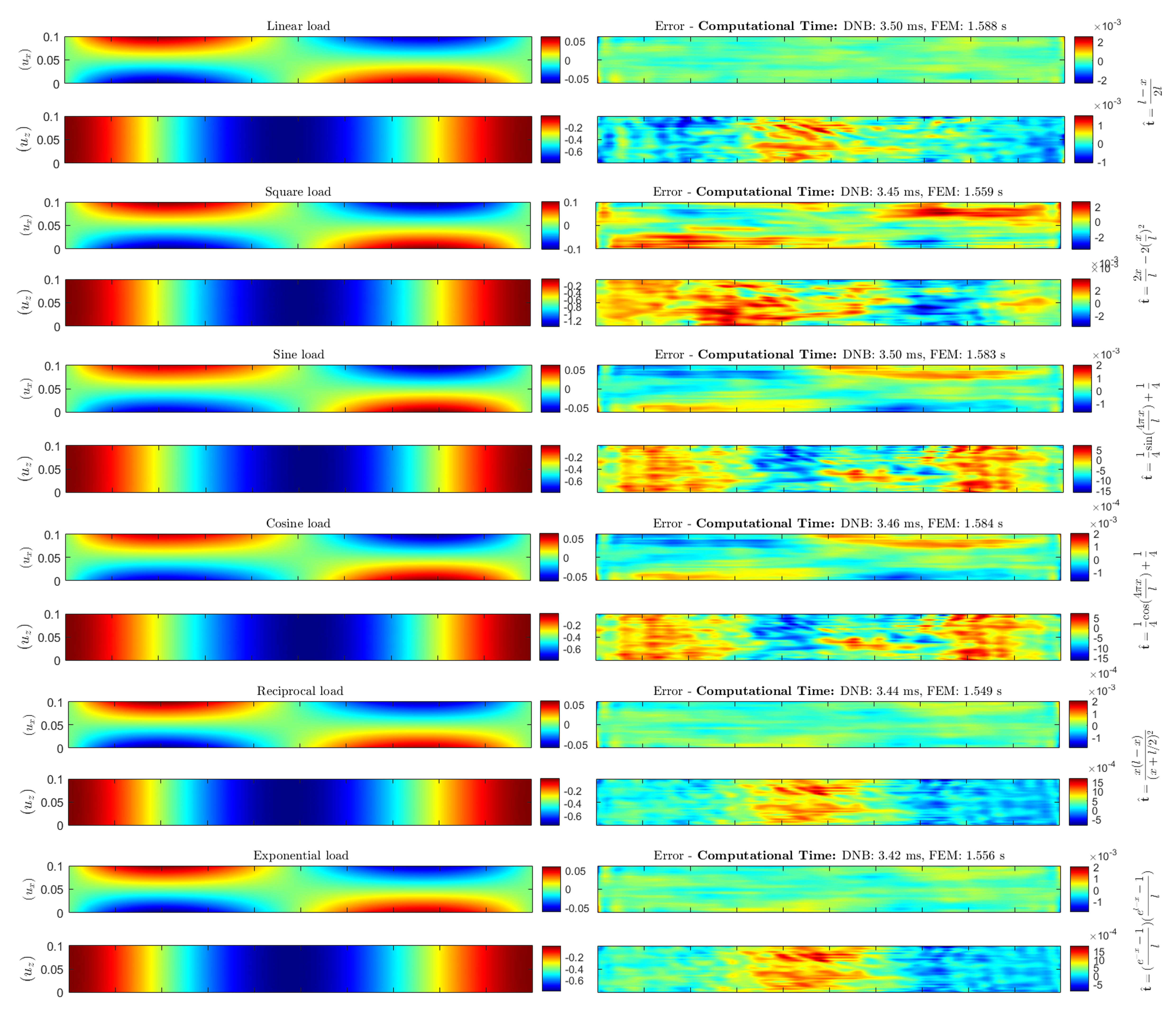}}
    \vspace{-30pt}
    \caption{Displacement \((\si{\centi\meter})\) field distribution for different traction function and symmetric material distribution}
    \label{fig:Displacement_field_x}
\end{figure}
where \(\mathcal{K}:\mathcal{A} \times \Theta_{\mathcal{K}} \mapsto \mathcal{L}(\mathcal{U}(D;\mathbb{R}^{d_v}), \mathcal{U}(D;\mathbb{R}^{d_v}) )\) denotes a mapping to bounded linear operators, parameterized by \(\phi \in \Theta_{\mathcal{K}}\), \(W:\mathbb{R}^{d_v} \mapsto \mathbb{R}^{d_v}\) is a linear transformation, and \(\sigma : \mathbb{R} \mapsto \mathbb{R}\) —applied component-wise— is a non-linear activation function. Moreover, the kernel integral operator mapping in Eq. \ref{eq:FourierLayer} is defined by 
\begin{equation}
\left( \mathcal{K}(a;\phi)\,v_t \right) (\chi) = \mathcal{F}^{-1}\left( \mathcal{F}(\kappa_\phi) \cdot \mathcal{F}(v_t) \right) (\chi), \quad \forall \chi \in \Omega \label{eq:FourierKernel}
\end{equation}
where \(\mathcal{F}\) denotes the Fourier transform of a function, and \(\mathcal{F}^{-1}\) represents its inverse, and \(\kappa_\phi\) is directly parameterized in Fourier space.

\begin{figure}[htb]
    \centering
    \makebox[\textwidth]{\includegraphics[width=1\textwidth]{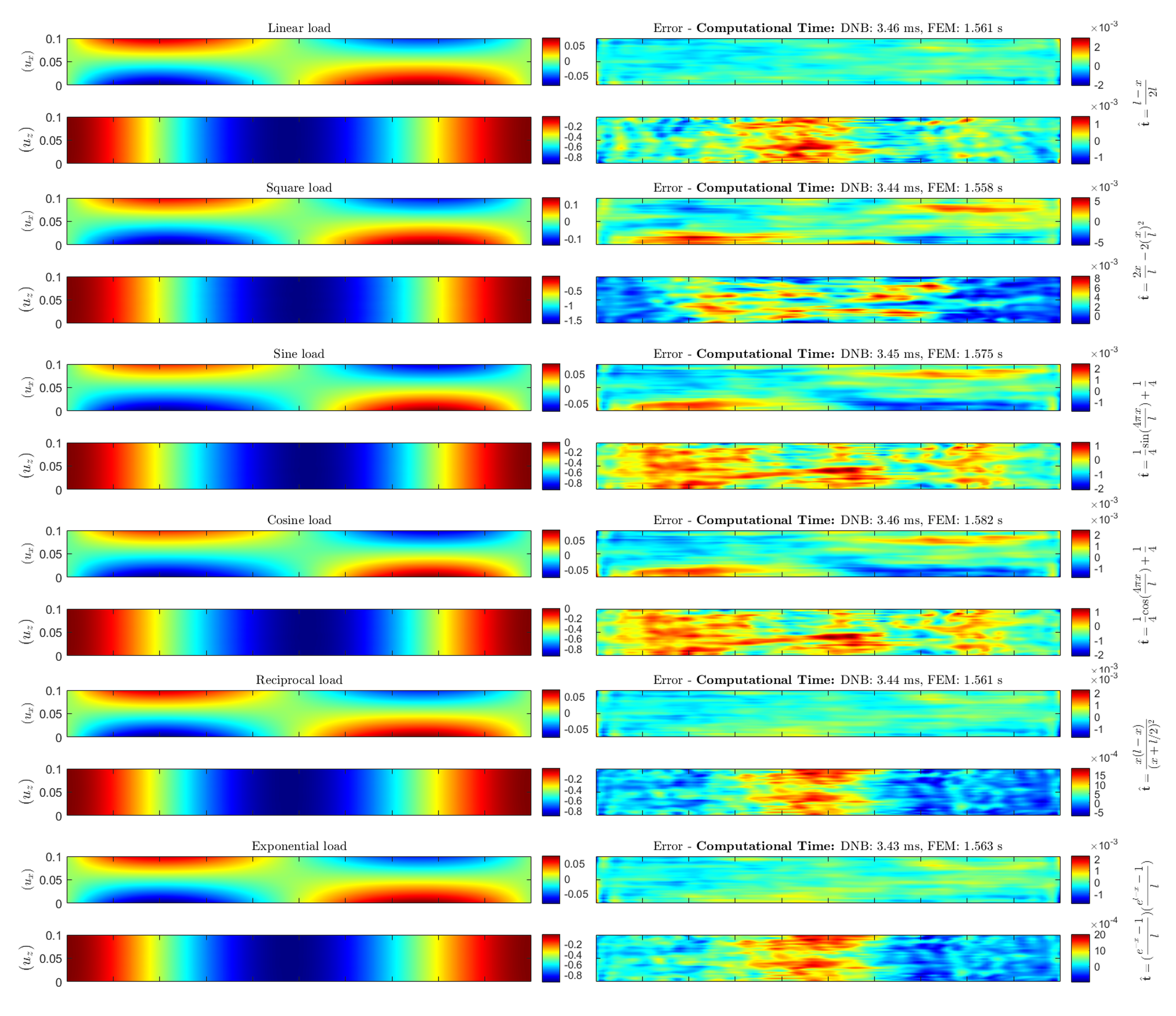}}
    \vspace{-30pt}
    \caption{Displacement \((\si{\centi\meter})\) field distribution for different traction function and asymmetric material distribution}
    \label{fig:Displacement_field_Z}
\end{figure}

The displacement field was computed using \(N_x \times N_z\) uniformly spaced points within the domain, with \(N_x = 128\) and \(N_z = 32\). Throughout all layers, the GELU (Gaussian Error Linear Unit) activation function was employed \cite{hendrycks2016gaussian}. The optimization process utilized the Adam optimizer, and the network underwent training on 4000 sets of GRF functions and was subsequently tested on 400 additional sets of GRF functions. The computational time taken for the training part is 337 s, but once the training is complete, solving the problem with any arbitrary function takes approximately 3 milliseconds, compared to 1.5 seconds for IGA. Finally, Fig. \ref{fig:loss_converge_FNO} displays the convergence of the loss function, while Figs. \ref{fig:Displacement_field_x}-\ref{fig:Displacement_field_Z} provides a visualization of the results obtained by employing the trained network to predict the displacement field for various traction functions and two material distributions.

\section{Parametric Analysis} \label{sec:ParametricAnalysis}
A comprehensive parametric analysis is conducted through a parametric study. The bending characteristics of porous beams are examined under different porosity coefficients, slenderness ratios, and various boundary conditions. 

\begin{figure}[htb]
    \centering
    \makebox[\textwidth][c]{\includegraphics[width=1.3\textwidth]{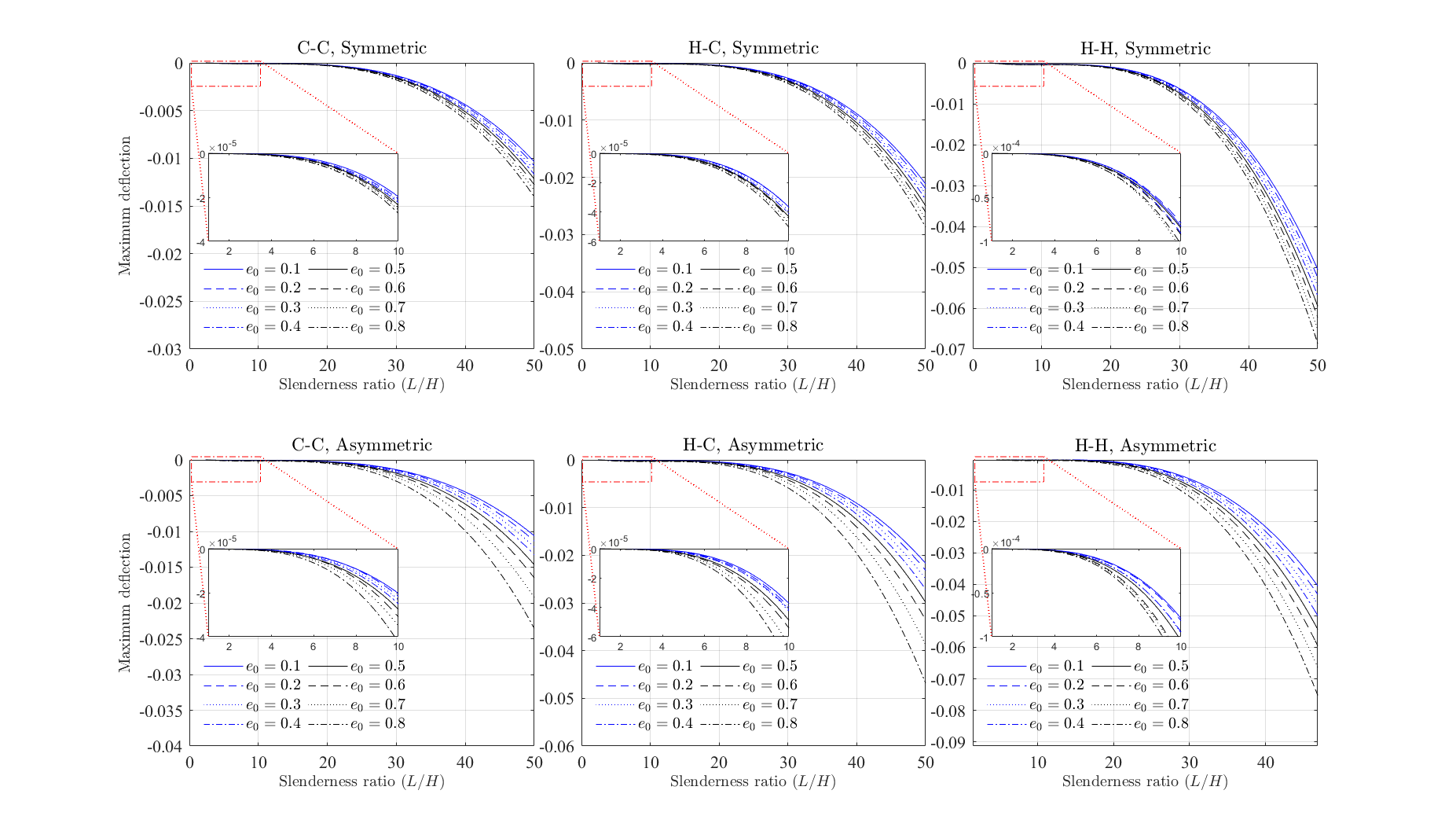}}
    \captionsetup{justification=centering,margin=2cm}
    \vspace{-30pt}
    \caption{Maximum deflection \((\si{\meter})\) under a uniformly distributed load: Effect of porosity coefficient and slenderness ratio}
    \label{fig:Maximum_deflection}
\end{figure}

Fig. \ref{fig:Maximum_deflection} depicts how the maximum deflection of a functionally graded porous beam under a uniformly distributed load is influenced by both the porosity coefficient and slenderness ratio. The graph showcases various boundary conditions and different porosity states. The trend is that increasing the porosity coefficient and slenderness ratio results in a greater deflection. Notably, beams with symmetric porosity distribution exhibit higher effective stiffness compared to those with asymmetric porosity distribution. Among the three considered boundary conditions (C-C, C-H, H-H), the H-H beam shows the largest deflection, while the C-C beam exhibits the smallest.

\begin{figure}[ht]
    \centering
    \makebox[\textwidth]{\includegraphics[width=1.15\textwidth]{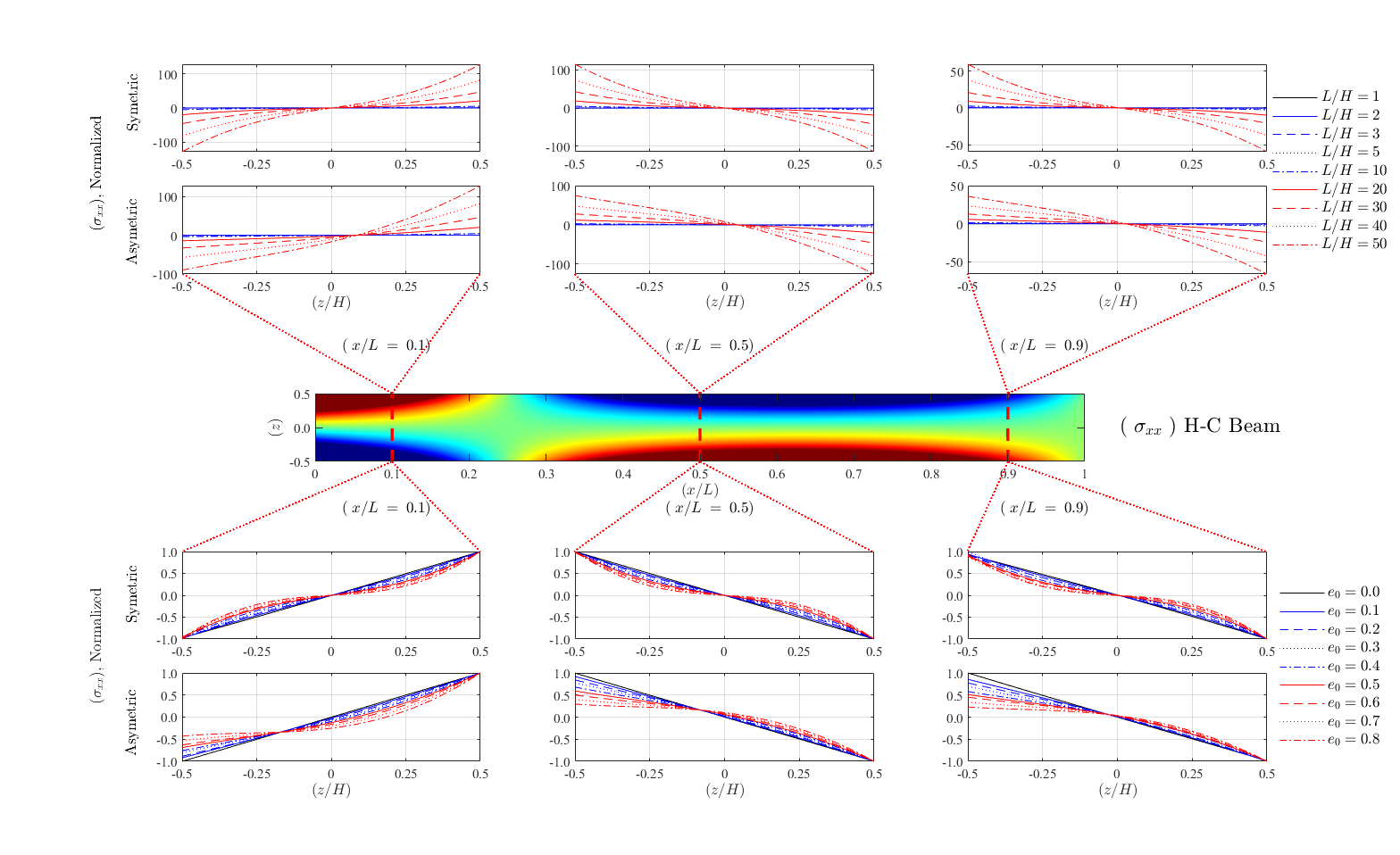}}
    \vspace{-40pt}
    \captionsetup{justification=centering,margin=0.5cm}
    \caption{Effect of porosity coefficient and slenderness ratio on the variation of dimensionless normal stress through the thickness for H-C beam under a distributed load at different \(x/L\).}
    \label{fig:Normal_Stress}
\end{figure}

Fig. \ref{fig:Normal_Stress} illustrates the impact of the porosity coefficient and slenderness ratio on the normalized normal stress across the thickness of an H-C beam subjected to a uniformly distributed load. Notably, the normal bending stress exhibits a linear variation along the thickness for a non-porous beam \(e_0 = 0\) while it displays a nonlinear pattern for functionally graded beams due to the non-uniform porosity distribution, resulting in a nonlinear gradient in material properties.

For beams with symmetric porosity distribution, a higher porosity coefficient corresponds to larger pore size and greater internal pore intensity, leading to lower local stiffness around the midplane. Consequently, this results in lower normal stress in the midplane region and higher stress on both the top and bottom surfaces. The normal bending stress is symmetric about the mid-plane for symmetric porosity distribution but asymmetric for asymmetric porosity distribution, where the pore size gradually increases from the top surface to the bottom surface. As a consequence, the maximum normal bending stress on the top surface is significantly larger than that on the bottom surface, with the difference increasing as \(e_0\) rises. Additionally, as expected, slender beams exhibit higher normal bending stress due to their weaker bending stiffness and larger deflections.

It is important to note that the aforementioned observations regarding normal bending stress, specifically for an H-H beam under a distributed load, are applicable to beams with various boundary and loading conditions, although those are not detailed here for conciseness.

To investigate different loading conditions, any well-behaved functions can be represented as a linear combination of sinusoidal functions through the Fourier transform:

\begin{equation}
f(x) = \frac{1}{\pi} \left[ \int_0^\infty A(k) \cos(kx) \, dk + \int_0^\infty B(k) \sin(kx) \, dk \right] \label{eq:Fourier}
\end{equation}
where ($k$) is the angular spatial frequency, $A(k)$ and $B(k)$ are weighting factors given by: 
\begin{equation}
A(k) = \int_{-\infty}^{+\infty} f(x') \cos(kx') \, dx'
\label{eq:Fourier_cosine}
\end{equation}
\begin{equation}
B(k) = \int_{-\infty}^{+\infty} f(x') \sin(kx') \, dx'
\label{eq:Fourier_sine}
\end{equation}

\begin{figure}[ht]
    \centering
    \makebox[\textwidth]{\includegraphics[width=1.15\textwidth]{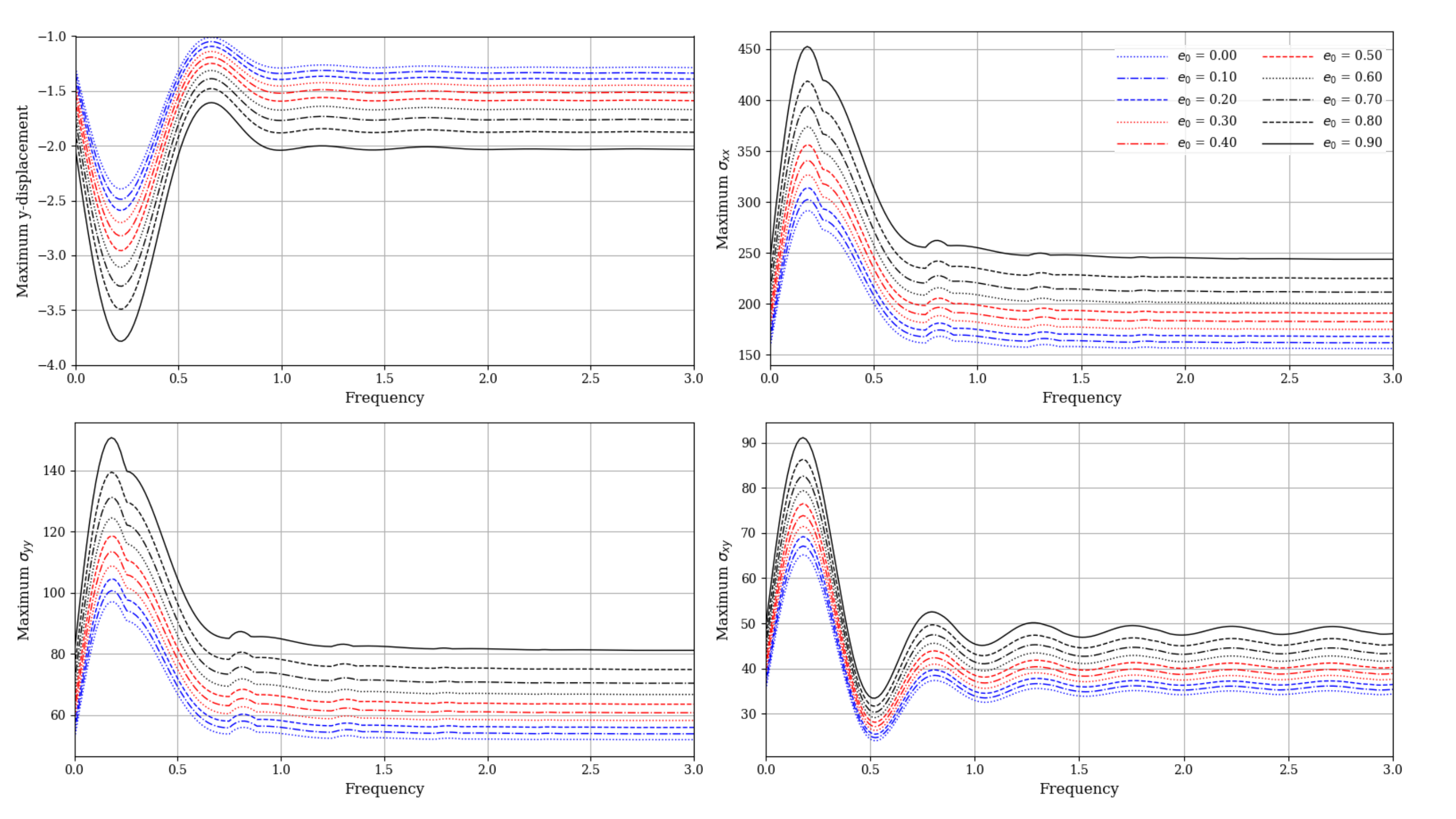}}
    \vspace{-20pt}
    \captionsetup{justification=centering,margin=0.5cm}
    \caption{Effect of various irregular loading conditions and porosity coefficient on the variation of maximum deflection (cm) and stress components (MPa) for C-C beam at different spatial frequencies}
    \label{fig:Irregular_loading}
\end{figure}

It is, therefore, sufficient to consider sine functions with varying frequencies. Thus, we examined the behavior of a functionally graded porous beam under distributed load with the form of:

\begin{equation}
t(x) = \frac{1}{2} \left( \sin(kx) + 1 \right)
\label{eq:irreqular_load}
\end{equation}
where \(k = 2\pi\nu\), and \(\nu\) represents the spatial frequency, which encompasses all positive real values. Fig. \ref{fig:Irregular_loading} summarizes the result of 2000 analyses and illustrates the maximum deflection and maximum stress components of the beam for various spatial frequencies and various porosity coefficients. The results show a similar trend across different porosity coefficients, with a peak occurring around a spatial frequency 0.25. Additionally, the beam's response for frequencies greater than 2 appears nearly identical.

\section{Conclusion} \label{sec:Conclusion}
The study investigated the application of scientific machine learning techniques, such as PINNs, DEM, and Neural Operators for functionally graded porous beam analysis for functionally graded porous beam analysis and unified them under one framework, named DeepNetBeam (DNB). This framework significantly improves the accuracy of beam analysis by minimizing reliance on the assumptions inherent to other beam theories. Through application in the analysis of FG porous beams, DNB showcases its versatility. Adopting SciML techniques ensures the efficient training of neural networks, yielding accurate results. Comparative analysis with exact solutions validates the efficacy of the methods, while the application of Neural Operators demonstrates a speed-up compared to traditional methods, notably in problems with arbitrary traction and material distribution. Parametric investigations on porous beams offer insights into the impact of parameters such as slenderness ratio and porosity coefficient on bending characteristics. 

In the DNB framework, three approaches have been implemented, each with its own advantages and disadvantages. PINNs offer the unique advantage of incorporating physics directly into the learning process, allowing for the inclusion of known physical laws and constraints. This makes them versatile and data-efficient. However, these strengths come with notable drawbacks. The computational cost of PINNs is a significant challenge, as solving partial differential equations during training is resource-intensive. Furthermore, successful application of PINNs requires a deep understanding of the underlying physics, and their performance can be highly sensitive to hyperparameter settings.

DEM excels by integrating energy-based physical principles during training, enhancing accuracy and reliability compared to purely data-driven methods. It is particularly advantageous over PINNs when dealing with second-order PDEs since only the first derivatives are required in DEM. However, DEM can introduce integration errors, which may compromise accuracy, particularly in scenarios where integration is complex. The precision of DEM's results largely depends on the methods used to compute numerical integrals within the loss function, which can be a source of error.

FNO is distinguished by its computational efficiency and resolution invariance. It can quickly solve not just specific PDEs but entire classes of PDEs, greatly increasing computational speed. Its resolution invariance allows it to be applied effectively across different spatial scales without sacrificing performance. However, FNO faces challenges with interpretability, as its parameterization can result in opaque outputs, making it difficult to understand the underlying mechanisms. Additionally, generalization is a concern; FNO may struggle to predict outcomes accurately for data outside the training distribution, limiting its use in diverse scenarios. Moreover, it requires a large dataset, which can be time-consuming and difficult to compile in practical situations.

Finally, this paper introduces a framework and demonstrates its applicability in structural analysis. By significantly reducing reliance on assumptions and showcasing efficient numerical solutions, it provides an opportunity for more accurate analysis of beam-like structures. The outcomes emphasize the potential of DNB in shaping the future of accurate and efficient structural analysis methodologies.

\section*{Declaration of Competing Interest}
The authors declare that they have no known competing financial interests or personal relationships that could have appeared to influence the work reported in this paper.

\section*{Acknowledgments}
The authors would like to acknowledge the support provided by the German Academic Exchange Service (DAAD) through a scholarship awarded to Mohammad Sadegh Eshaghi during the course of this research.

\printbibliography
\end{document}